%% file: alignment.tex
\documentclass[10pt,twocolumn,letterpaper]{article}

\usepackage{conf}
\usepackage{times}
\usepackage{epsfig}
\usepackage{graphicx}
\usepackage{amsmath}
\usepackage{amssymb}
\usepackage{dblfloatfix}

\usepackage{color}

\usepackage[pagebackref=true,breaklinks=true,letterpaper=true,colorlinks,bookmarks=false]{hyperref}

\newcommand\Black{\color{black}}  

\newcommand{\Id}{\mathrm{Id}}
\newcommand\R{\mathbb{R}}

\newcommand \E{\mathop{\mathbb{E}}}
\newcommand \x {\mathbf{x}}

\newcommand \w {\mathbf{w}}

\newcommand{\vv}{\mathbf{v}}
\newcommand \deriv[2]{\frac{\partial{#1}}{\partial{#2}}}

\newcommand \Distr {\mathcal{D}}
\newcommand\?{\mkern-1.5mu}   
\newcommand\GT{{\tiny\mbox{GT}}}
\newcommand\Laplace{\mathcal{4}}

\graphicspath{{allfigs/}}

\begin{document}

\title{Coarse to fine non-rigid registration:\\
  a chain of scale-specific neural networks for multimodal image alignment\\
  with application to remote sensing}

\author{
Armand Zampieri${}^{1}$, Guillaume Charpiat${}^{2}$, Yuliya Tarabalka${}^{1}$\\
\emph{${ }^{1}$ Titane team, INRIA Sophia-Antipolis}\\
\emph{${ }^{2}$ TAU team, LRI, INRIA Saclay, Universit\'e Paris-Sud}\\
{\tt\small \{guillaume.charpiat,yuliya.tarabalka\}@inria.fr}
}

\maketitle

\emph{This project started early 2017 and this paper was submitted for publication in November 2017.\\ \vspace{2mm}}

\begin{abstract}
  We tackle here the problem of multimodal image non-rigid registration, which is of prime importance in remote sensing and medical imaging. The difficulties encountered by classical registration approaches include feature design and slow optimization by gradient descent. By analyzing these methods, we note the significance of the notion of scale. We design easy-to-train, fully-convolutional neural networks able to learn scale-specific features. Once chained appropriately, they perform global registration in linear time, getting rid of gradient descent schemes by predicting directly the deformation.

  We show their performance in terms of quality and speed through various tasks of remote sensing multimodal image alignment. In particular, we are able to register correctly cadastral maps of buildings as well as road polylines onto RGB images, and outperform current keypoint matching methods.
\end{abstract}


\input{intro.tex}  

\input{analysis.tex} 
\input{modelisation.tex} 


\input{experiments.tex} 

\input{discussion.tex}  

{\small
\bibliographystyle{ieee}
\bibliography{alignment}
}

\appendix
\section{Appendix}

\input{annexe.tex}


\end{document}

%% file: intro.tex


\begin{figure}[h]
\begin{center}
  \includegraphics[width=0.47\linewidth]{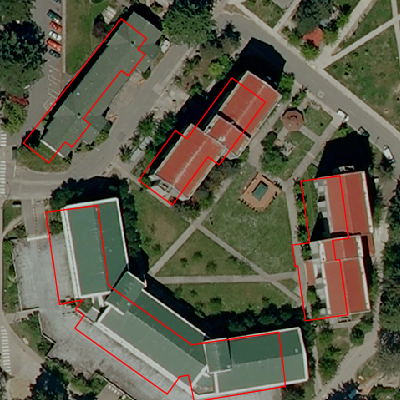}
   \includegraphics[width=0.47\linewidth]{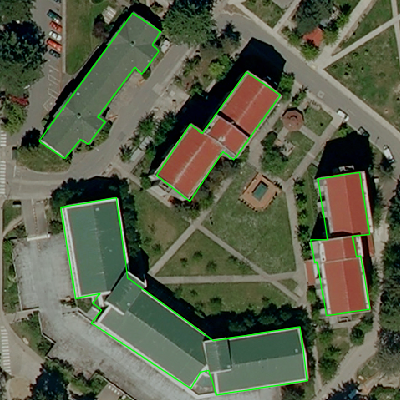}
   \caption{\label{fig:teaser}\textbf{Multimodal matching.} We align aerial images with cadastral images. Left: original misregistered images, right: after our realignment.}
\end{center}
\end{figure}

\section{Introduction}

Image alignment, also named \emph{non-rigid registration}, is the task of finding a correspondence field between two given images, \ie a deformation which, when applied to the first image, warps it to the second one. Such warpings can prove useful in many situations: to transfer information between several images (for instance, from a template image with labeled parts), to compare the appearance of similar parts (as pixel intensity comparison makes sense only after alignment), or to estimate spatial changes (to monitor the evolution of a tumor given a sequence of scans of the same patient over time, for instance). Image alignment has thus been a predominant topic in fields such as medical imaging 
or remote sensing \cite{sotiras2013deformable,med}.

\subsection{Remote sensing \& Image alignment}
\label{sec:intro_remote}
In remote sensing, images of the Earth can be acquired through different types of sensors, in the visible spectrum (RGB) or not (hyperspectral images), from satellites or planes, with various spatial precision (from cm to km range). The analysis of these images allows the monitoring of ecosystems (plants~\cite{Hansen850}, animals~\cite{verdie_eccv12}, soils...) and their evolution (ice melting,
drought monitoring, natural disasters and associated help planning),
    urban growth, as well as the automatic creation of maps \cite{mattyus2016hd} or more generally digitizing the Earth.

    However, the geographic localization of pixels in remote sensing images is limited by a number of factors, such as the positioning precision and the effect of the relief on non-vertical points of view.
The deformation of these images is significant: for instance, in OpenStreetMap~\cite{haklay08}, objects may be shifted by 8 meters (which is far above the required precision of maps for autonomous driving), which means an error displacement of more than 20 pixels for a 30 cm/pixel resolution.



These deformations prevent a proper exploitation of such data. For instance, let us consider the task of finding buildings and roads in a remote sensing image. While ground truth is actually available in considerable amounts, such as in OpenStreetMap (OSM) based on cadastral information, which gives coordinates (latitude and longitude) of each building corner, this hand-made ground truth is often inaccurate, because of human mistakes.
Thus it is not possible to learn from it, as remote sensing images are not properly aligned to it and objects might even not overlap. This is a severe issue for the remote sensing field in the era of big data and machine learning. Many works have been focusing on this problem \cite{segm}, from the use of relief knowledge
to dedicated hand-designed alignment algorithms.
Another approach worth mentioning is to train coarsely on the datasets available and fine-tune on small better-hand-aligned datasets \cite{DBLP:journals/corr/MaggioriTCA16}. We will here tackle the problem of non-rigid alignment directly.

\subsection{Classical approaches for non-rigid registration}

\paragraph{Tasks} Image registration deals with images either of the same modality (same sensor), or not. When of the same modality, the task is typically to align different but similar objects (\eg, faces \cite{charpiat-keriven-etal:05c} or organs of different people \cite{hermosillo2002variational}),
or to align the same object but taken at different times (as in the tumor monitoring example). On the other side, multi-modal registration deals with aligning images usually of the same object but taken with different sensors, which capture different physical properties, at possibly different resolutions. For instance in medical imaging, MR and CT scans capture the density of water and of matter respectively, while in remote sensing RGB and hyperspectral data capture information from different light frequencies (infrared, etc.). In our case of study, we will focus on the alignment of RGB remote sensing images with cadastral maps, which are vector-format images with polygonal representations of all buildings and roads, hand-made by local authorities, map makers or OpenStreetMap users, as in Figures \ref{fig:teaser} and \ref{fig:dataset}.\\

Whether mono-modal or multi-modal, image registration faces two challenges: first, to describe locally image data, and then, to match points with similar description, in a spatially coherent manner. Historically, two main classical approaches have emerged.

\paragraph{Matching key-points} The first one consists in sampling a few key-points from each image (\eg with Harris corner detection), in describing them locally (with SIFT, SURF, HOG descriptors...) \cite{sift,hog}, in matching these points \cite{mikolajczyk2004scale} and then interpolating to the rest of the image.
The question is then how to design proper sampling criteria, descriptors, and matching algorithm.
In the multi-modal case, one would also have to design or learn the correspondence between the descriptors of the two modalities.
Note that high-precision registration requires a dense sampling, as well as consequently finer descriptors, which naturally leads to the second approach.

\paragraph{Estimating a deformation field by gradient descent}
\label{sec:classic_grad}
The second approach, particularly popular in medical imaging, consists in estimating a dense deformation field from one image to the other one
\cite{beg2005computing,glaunes2004diffeomorphic,michor2007metric,kendall1989survey,hermosillo2002variational}.
One of its advantages over the first approach is to be able to model objects, to make use of shape statistics, etc.
The warping is modeled as a smooth vector field $\phi$, mapping one image domain onto the other one.
Given two images $I_1$ and $I_2$, a criterion $C(I_1 \!\circ\? \phi, \, I_2)$ is defined, to express the similarity between the warped image $I_1 \!\circ\? \phi$ and the target $I_2$, and is optimized with respect to $\phi$ by gradient descent. Selecting a suitable similarity criterion $C$ is crucial, as well as designing carefully the gradient descent, as we will detail in section \ref{sec:scale}.

\subsection{The new paradigm: neural networks} 

The difficulty to design or pick particular local descriptors or matching criteria among many possibilities is the trait of computer vision problems where the introduction of neural networks can prove useful. The question is how. Machine learning techniques have already been explored to learn similarity measures between different imaging modalities \cite{comp}, for instance using kernel methods to register MR and CT brain scans \cite{lee2009learning}, but without tackling the question of scale. We will aim at designing a system able to learn scale-specific and modality-specific features, and able to perform multimodal image registration densely and swiftly, without the use of any iterative process such as gradient descent which hampers classical approaches.
Our contributions are thus:.
\begin{itemize}\itemsep-0.2em 
\item a swift system to register images densely
\item learning features to register images of different modalities
\item learning scale-specific features and managing scales
\item designing a (relatively small) neural network to do this in an end-to-end fashion
\item aligning remote sensing images with cadastral maps (buildings and roads)
\item providing a long-awaited tool to create large-scale benchmarks in remote sensing.
\end{itemize}
We first analyze the problems related to scale when aligning images, in order to design a suitable neural network architecture. We show results on benchmarks and present additional experiments to show the flexibility of the approach.


%% file: analysis.tex

\section{Analysis of the gradient descent framework}

\noindent In order to analyze issues that arise when aligning images, let us first consider the case of mono-modal registration, for simplicity. Keeping the notations from section \ref{sec:classic_grad}, we pursue the quest for a reasonable criterion $C(I_1 \!\circ\? \phi, \, I_2)$ to optimize by gradient descent to estimate the deformation~$\phi$.

\label{sec:scale}
\subsection{A basic example}


Too local quantities such as the pixellic intensity difference $C(I_1 \!\circ\? \phi, \, I_2) = \left\| I_1 \!\circ\? \phi - I_2 \right\|^2_{L^2}$ would create many local minima and get the gradient descent stuck very fast. Indeed, if as a toy example one considers $I_1$ and $I_2$ to be two white images with a unique black dot\footnote{In the continuous setting, consider a smooth compact-support bump.}
at different locations $\x_1$ and $\x_2$ respectively, the derivative of $C(I_1 \!\circ\? \phi, \, I_2)$ with respect to $\phi$ will never involve quantities based on these two points at the same time, which prevents them from being influenced by each other:
$$\deriv{C(I_1 \!\circ\? \phi, \, I_2)}{\phi}(\x) = 2 \big(I_1 \!\circ\? \phi(\x) - I_2(\x)\big)\, \left(\nabla_{\!\x} I_1\right)_{(\phi(\x))}$$
is 0 at all points $\x \neq \x_1$,  and at $\x_1$ the deformation $\phi$ (initialized to the identity) evolves to make it disappear from the cost $C$ by shrinking the image around. Thus the derivative of the similarity cost $C$ with respect to the deformation $\phi$ does not convey any information pushing $\x_1$ towards $\x_2$, but on the contrary will make the descent gradient stuck in this (very poor) local minimum.

Instead of the intensity $I(\x)$, one might want to consider other local, higher-level features $L(I)(\x)$ such as edge detectors, in order to grasp more meaningful information, and thus minimize a criterion for instance of the form:
\begin{equation}
\label{eq:criterion_feature}
C(I_1 \!\circ\? \phi, \, I_2) = \left\| L(I_1 \!\circ\? \phi) - L(I_2) \right\|^2_{L^2}.
\end{equation}


\subsection{Neighborhood size}
\label{sec:neighsize}
The solution consists in considering local descriptors involving larger spatial neighborhoods, wide enough so that the image domains involved in the computations of $L(I_1 \!\circ\? \phi)(\x_1)$ and $L(I_2)(\x_2)$
for two points $\x_1$ and $\x_2$ to be matched
overlap significantly.
For instance, the computation of the Canny edge detector is performed over a truncated Gaussian neighborhood, whose size is pre-defined by the standard deviation parameter $\sigma$. Another example is the local-cross correlation, which compares the local variations of the intensity around $\x_1$ and $\x_2$ within a neighborhood of pre-defined size \cite{hermosillo2002variational}. Another famous example is the mutual information between the histograms of the intensity within a certain window with pre-defined size.

\subsection{Adapting the scale}
In all these cases, the neighborhood size is particularly important: if too small, the gradient descent will get stuck in a poor local minimum, while if too large, the image details might be lost, preventing fine registration.
What is actually needed is this neighborhood size to be of the same order of magnitude as the displacement to be found. As this displacement is unknown, the neighborhood size needs to be wide enough during the first gradient steps (possibly covering the full image), and has to decrease with time, for the registration to be able to get finer and finally reach pixellic precision. Controlling the speed of this decrease is a difficult matter, leading to slow optimization. Moreover, the performance of the descriptors may depend on the scale, and different descriptors might need to be chosen for the coarse initial registration than for the finest final one.
In plus of the difficult task of designing \cite{yu2008fast,ye2014local} or learning \cite{lee2009learning} relevant descriptors $L_s$ for each scale, this raises another issue, that is that the criterion $C_s$ to optimize
\begin{equation}
\label{eq:scale_dep}
C_s(I_1 \!\circ\? \phi, \, I_2) = \left\| L_s(I_1 \!\circ\? \phi) - L_s(I_2) \right\|^2_{L^2}
\end{equation}
now depends on the current neighborhood size $s(t)$, which is itself time-dependent, and thus the optimized criterion $C_{s(t)}$ might increase when the descriptor $L_{s(t)}$ evolves: the optimization process is then not a gradient descent anymore.

One might think of scale-invariant descriptors such as SIFT, however the issue is not just to adapt the scale to a particular location within an image, but to adapt it to the amplitude of the deformation that remains to be done to be matched to the \emph{other} image.

\subsection{Multi-resolution viewpoint}
Another point of view on this scale-increasing process
is to consider that the descriptors and optimization process remain the same at all scales, but that the \emph{resolution} of the image is increasing. The algorithm is then a loop over successive resolutions \cite{hermosillo2002variational,charpiat-keriven-etal:05c}, starting from a low-resolution version of the image, waiting for convergence of the gradient descent, then upsampling the deformation field found to a higher resolution version of the image, and iterating until the original resolution is reached. The limitation is then that the same descriptor has to be used for each resolution, and, as previously, that the convergence of a gradient descent has to be reached at each scale, leading to slow optimization.
A different approach consists in dealing with all scales simultaneously by considering a multi-scale parameterisation of the deformation \cite{schnabel2001generic}. However, the same local minimum problem would be encountered if implemented naively; heuristics then need to be used to estimate at which scale the optimization has currently to be performed locally.

\subsection{Keeping deformations smooth}
Deformations are usually modeled as diffeomorphisms
\cite{beg2005computing,glaunes2004diffeomorphic,kendall1989survey,hermosillo2002variational},
\ie smooth one-to-one vector fields, in order to avoid deleting image parts. The smoothness is controlled by an additional criterion to optimize, quantifying the regularity of the deformation $\phi$, such as its Sobolev norm (penalizing fast variations).
As in any machine learning technique, this regularity term sets a prior over the space of possible functions (here, deformations), preventing overfitting (here, spatial noise). But once again, the smoothness level required should depend on the scale, \eg prioritizing global translations and rotations at first, while allowing very local moves when converging. This can be handled by suitable metrics on instantaneous deformations \cite{charpiat-pons-etal:07,sundaramoorthi2008coarse}; yet in practice these metrics tend to slow down convergence by over-smoothing gradients $\nabla_{\!\phi}\, C$ at finest scales.

%% file: modelisation.tex


\section{Introducing neural networks}

\subsection{Learning iterative processes}
As neural networks have proved useful to replace hand-designed features for various tasks in the literature recently, and convolutional ones (CNN) in particular in computer vision, one could think, for mono-modal image alignment, of training a CNN in the Siamese network setup
\cite{bromley1994signature,chopra2005learning},
in order to learn a relevant distance between image patches. The multi-modal version of this would consist in training two CNN (one per modality) with same output size, in computing the Euclidean norm of the difference of their outputs as a dissimilarity measure, and in using that quantity within a standard non-rigid alignment algorithm, such as a gradient descent over (\ref{eq:criterion_feature}). For training, this would however require to be able to differentiate the result of this iterative alignment process with respect to the features. This is not realistic, given the varying, usually large number of steps required for typical alignment tasks. A similar approach was nonetheless successfully used in \cite{DBLP:journals/corr/MaggioriTCA16}, for the simpler task of correcting blurry segmentation maps, sharpening them and relying on image edges. For this, a partial differential equation (PDE) was mimicked with a recurrent network, and the number of steps applying this PDE was pre-defined to a small value (5), sufficient for that particular problem. In the same spirit, for image denoising, in \cite{meinhardt2017learning,wang2016proximal} the proximal operator used during an iterative optimization process is modeled by a neural network and learned. There is to our knowledge no neural network-based work for dense non-rigid alignment yet. In \cite{rs9060586}, the Siamese network idea is used, but for matching only very few points. It is also worth noting that, much earlier, in \cite{lee2009learning}, a similarity criterion between different modalities was learned, with a kernel method, but for rigid registration only.

\subsection{A more direct approach}

As seen in the previous sections, aligning images thanks to a gradient descent over the deformation $\phi$ has the following drawbacks: it is slow because of the need to ensure convergence at each scale, it is actually not a real gradient descent if descriptors are scale-dependent, and it induces a long backpropagation chain when learning the descriptors. To get rid of this iterative process, we propose to predict directly its final result at convergence. That is, given images $I_1$ and $I_2$, to predict directly the optimal deformation $\phi$ so that $I_1\!\circ\?\phi$ and $I_2$ are aligned. Also, instead of proceeding in two steps: first learning the features $L$ required to define the criterion $C$ in (\ref{eq:criterion_feature}), then finding the deformation $\phi$ minimizing $C$, we propose to directly learn the deformation as in a standard machine learning setup, that is, from examples. Given a training dataset of input pairs $P = (I_1, I_2)$ together with the expected associated output $\phi_P$, we aim at learning the function $P \mapsto \phi_P$.

\subsection{Machine learning setting}
\begin{figure}[t]
\begin{center}
   \includegraphics[width=0.5\linewidth]{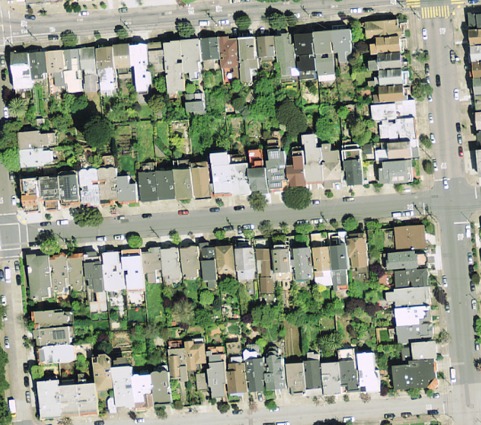}
   \includegraphics[width=0.488\linewidth]{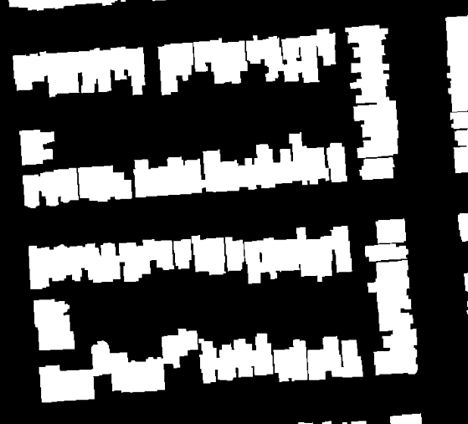}
   \caption{\label{fig:dataset} {\bf Multimodal pair} of already satisfyingly aligned images, from the database. Left: aerial RGB image, right: binary vector-format cadastral image (buildings are shown in white).
}
\end{center}
\end{figure}

\paragraph{Training set} We first consider the task of aligning geolocalized aerial RGB images with binary maps from OpenStreetMap indicating building locations. 
As explained in section \ref{sec:intro_remote}, the matching is usually imperfect. Creating the deformation ground truth by manually performing the warpings would be too time-consuming. Instead, we extract image pairs which visually look already well aligned, as in Figure \ref{fig:dataset}. This way we obtain a dataset composed of 5000x5000 image pairs (aerial RGB image, binary vector-format building map) at resolution $0.3$m/pixel, for which the deformation $\phi$ to be found is the identity.

We generate an artificial training set by applying random deformations to the cadastral vectorial maps, moving accordingly the corners of the polygons it contains, and then generating the new binary maps by rasterization. We thus obtain a training set of pairs of non-registered images, with known deformations.
As typical deformations in reality are smooth, we model our family of random deformations as: a global translation $\vv_0$ taken uniformly within a certain range $[-r,+r]^2$, plus a mixture of Gaussian functions with random shifts $\vv_i$, centers $\x_i$ and covariance matrices $S_i$:
\begin{equation}
\phi(\x) \;=\; \vv_0 \,+\, \sum_{i=1}^{n} \vv_i\; e^{-(\x -\x_i) \,S_i\, (\x - \x_i)}
\end{equation}
with uniformly random $\vv_i,\, S_i,\, \x_i$ within suitable pre-defined ranges ($S_i$ being symmetric positive definite). This way, we can drastically augment the dataset by applying arbitrarily many random deformations to the initially well-aligned images.

\paragraph{Optimization criterion} The loss considered is simply the Euclidean norm of the prediction error:
$$C(\w) = \!\!\!\!\E_{(I_1, I_2, \phi_\GT) \in  \Distr}\!\left[\sum_{\x \in \Omega(I_2)} \left\|\, \widehat\phi_{(\w)(I_1, I_2)}(\x) - \phi_\GT(\x) \,\right\|^2_2 \right]$$
\ie the expectation, over the ground truth dataset $\Distr$ of triplet examples (RGB image $I_1$, cadastral image $I_2$, associated deformation $\phi_\GT$), of the sum, over all pixels $\x$ in the image domain $\Omega(I_2)$, of the norm of the difference between the ground truth deformation $\phi_\GT(\x)$ and the one predicted $\widehat\phi_{(\w)(I_1, I_2)}(\x)$ for the pair of images $(I_1,I_2)$ given model parameters $\w$ (\ie the neural network weights).

In order to make sure that predictions are smooth, we also consider for each pixel a penalty over the norm of the (spatial) Laplacian of the deformation $\widehat\phi$:
\begin{equation}
\label{eq:laplacien}
+ \left\| \,\Laplace \widehat\phi_{(\w)(I_1, I_2)}\, (\x) \,\right\|^2_2
\end{equation}
which penalizes all but affine warpings.
In practice in the discrete setting this sum is the deviation of $\widehat\phi(\x)$ from the average over the 4 neighboring pixels:
$+\left\|\,\widehat\phi(\x) -  \frac{1}{4}\sum_{\x'\sim\x}\widehat\phi(\x')\,\right\|_2^2.$

\subsection{A first try}
\label{sec:first_try}
We first produce a training set typical of real deformations by picking
a realistic range $r = \pm 20$ pixels of deformation amplitudes.
We consider a fully-convolutional neural network, consisting of two convolutional networks (one for each input image $I_i$), whose final outputs are concatenated and sent to more convolutional layers. The last layer has two features, \ie emits two real values per pixel, which are interpreted as $\widehat\phi(\x)$.
In our experiments, such a network does not succeed in learning deformations: it constantly outputs  $\widehat\phi(\x) = (0,0)\; \forall \x$, which is the best constant value for our loss, \ie the best answer one can make when not understanding the link between the input $(I_1,I_2)$ and the output $\phi$ for a quadratic loss: the average expected answer $\E_{(I_1, I_2, \phi_\GT) \in  \Distr}\left[\phi\right]$, which is $(0,0)$ in our case.

We also tried changing representation by predicting bin probabilities $p\big(\Phi_x(\x) \in [a,a+1]\big),\; p\big(\Phi_y(\x) \in [b,b+1]\big)$ for each integer $-r \leqslant a,b < r$, by outputting 2 vectors of $2r$ real values per pixel, but this lead to the same result.



\subsection{Dealing with a single scale}
\label{sec:singlescale}

The task in section \ref{sec:first_try} is indeed too hard: the network needs to develop local descriptors at all scales to capture all information, and is asked to perform a fine matching with $(2r)^2 \simeq 1700$ possibilities for each pixel $\x$.

This task can be drastically simplified though, by requiring the network to perform the alignment \emph{at one scale $s$ only}. By this, we mean:\smallskip

\noindent \fbox{\parbox{0.97\linewidth}{\vspace{0.6mm}
    \underline{\bf Task at scale $s$:} $\;$ \emph{Solve the alignment problem for the image pair $(I_1, I_2)$, with a precision required of $\pm 2^s$ pixels, under the assumption that the amplitude of the registration to be found is not larger than $2^{s+1}$ pixels.}
    \vspace{0.2mm}}}\medskip

\noindent For instance, at scale $s=0$, the task is to search for a pixelwise precise registration ($\pm1$ pixel) on a dataset prepared as previously but with amplitude $r = 2^{s+1} = 2$. As a first approximate test, we train the same network as described earlier, with $r=2$, \ie each of the 2 coordinates of $\phi(\x)$ take value in $[-2,2]$, and we consider a prediction $\widehat\phi(\x)$ to be correct if in the same unit-sized bin as $\phi(\x)$.
Without tuning the architecture or the optimization method, we obtain, from the first training run, about $90\%$ of accuracy, to be compared to $\sim 6\%$ for a random guess.

Thus, it is feasible, and easy, to extract information when specifying the scale. Intuitively, the search space is much smaller; in the light of section \ref{sec:neighsize}, the descriptor receptive field required for such a $\pm 1$ pixel task is just of radius 1. And indeed, in the classical framework for mono-modal registration, a feature as simple as the image intensity would define a sufficiently good criterion (\ref{eq:criterion_feature}), as the associated gradient step involves the comparison to the next pixel (through $\nabla_{\!\x} I_1$). Note that such a simple intensity-based criterion would not be expected to perform more, \eg find deformations of amplitude $r \geqslant 2$ pixels in the general case (textures).

\paragraph{Designing a suitable neural network architecture}  We now propose better architectures to solve that alignment task at scale $s=0$. 
We need a fully-convolutional architecture since the output is a 2-channel image of the same size as the input, and we need to go across several scales in order to understand which kind of object part each pixel belongs to, in each modality. High-level features require a wide receptive field, and are usually obtained after several pooling layers, in a pyramidally shaped network. The output needs however to be of the same resolution as the input, which leads to autoencoder-like shapes. In order to keep all low-level information until the end, and not to lose precision, we link same-resolution layers together, thus obtaining a U-net-like network \cite{U-net} (developed for medical image segmentation). As the 2 input images are not registered, and to get modality-specific features, we build 2 separate convolutional pyramids, one for each modality (in a similar fashion as networks for stereo matching \cite{stereo}), but concatenate their activities once per scale to feed the U-net.
The architecture is summarized in Figure \ref{fig:net}. The network is trained successfully to solve the $s=0$ task as explained previously.

\subsection{A chain of scale-specific neural networks}
\label{sec:fullnet}

We now solve the general alignment task very simply:\smallskip

\noindent \fbox{\parbox{0.97\linewidth}{\vspace{0.6mm}
    \underline{\bf Solution for task at scale $s$:} $\;$ \emph{Downsample the images by a factor $2^s$; solve the alignment task at scale $0$ for these reduced images, and upsample the result with the same factor.}
    \vspace{0.2mm}}}\medskip

\noindent \fbox{\parbox{0.97\linewidth}{\vspace{0.6mm}
    \underline{\bf Full alignment algorithm:} $\;$ \emph{Given an image pair $(I_1, I_2)$ of width $w$, iteratively solve the alignment task at scale $s$, from $s=\log_2 w$ until $s=0$.}
    \vspace{0.2mm}}}\medskip

One can choose to use the same network for all scales, or different ones if we expect specific features at each scale, as in remote sensing or medical imaging.

The full processing chain is shown in Figure \ref{fig:chain}. Note a certain global similarity with ResNet \cite{DBLP:journals/corr/HeZRS15}, in that we have a chain of consecutive scale-specific blocks, each of which refining the previously-estimated deformation, not by adding to it
but by diffeomorphism composition: $\phi_{s-1} = \phi_s \circ \big(\Id + f(I_1 \!\circ \phi_s, I_2 \!\circ \phi_s)\big)$. Another difference with ResNet is that we train each scale-specific block independently, which is much easier than training the whole chain at once.

Note that the overall complexity of an alignment is very low, linear in the image size. Indeed, for a given image with $n$ pixels, a similar convolutional architecture is applied to all reduced versions by factors $2^s$, of size $2^{-s}\times 2^{-s} n$ pixels, leading to a total cost of $n (1 + \frac{1}{4} + \frac{1}{16} + \frac{1}{64} + \dots ) K <~\frac{4}{3} n K$ where $K$ is the constant per-pixel convolutional cost. This is to be compared with the classical gradient descent based approaches of unknown convergence duration, and with the classical multi-resolution approaches with gradient descents at each scale.

Note also some similarity with recent work on optical flow \cite{ilg2016flownet}, consisting in an arrangement of 3 different scale-related blocks, though monomodal, not principled from a scale analysis and without scale-specific training.\\


\noindent We will also check the following variations:
\begin{itemize}\itemsep0em
\item ``\emph{fast}'': replace all scale-specific blocks with the same $s=2$-specific block, to see how well features generalize across scales; the output quality decreases slightly but remains honorable.
\item ``\emph{accurate}'': apply the network on symmetrised and rotated versions of the input images, and average the result over these 8 tests. This ensures rotation invariance and improves the result.
\end{itemize}

%% file: experiments.tex

\section{Experiments}

\begin{figure}[t]
\begin{center}
   \includegraphics[width=0.99\linewidth]{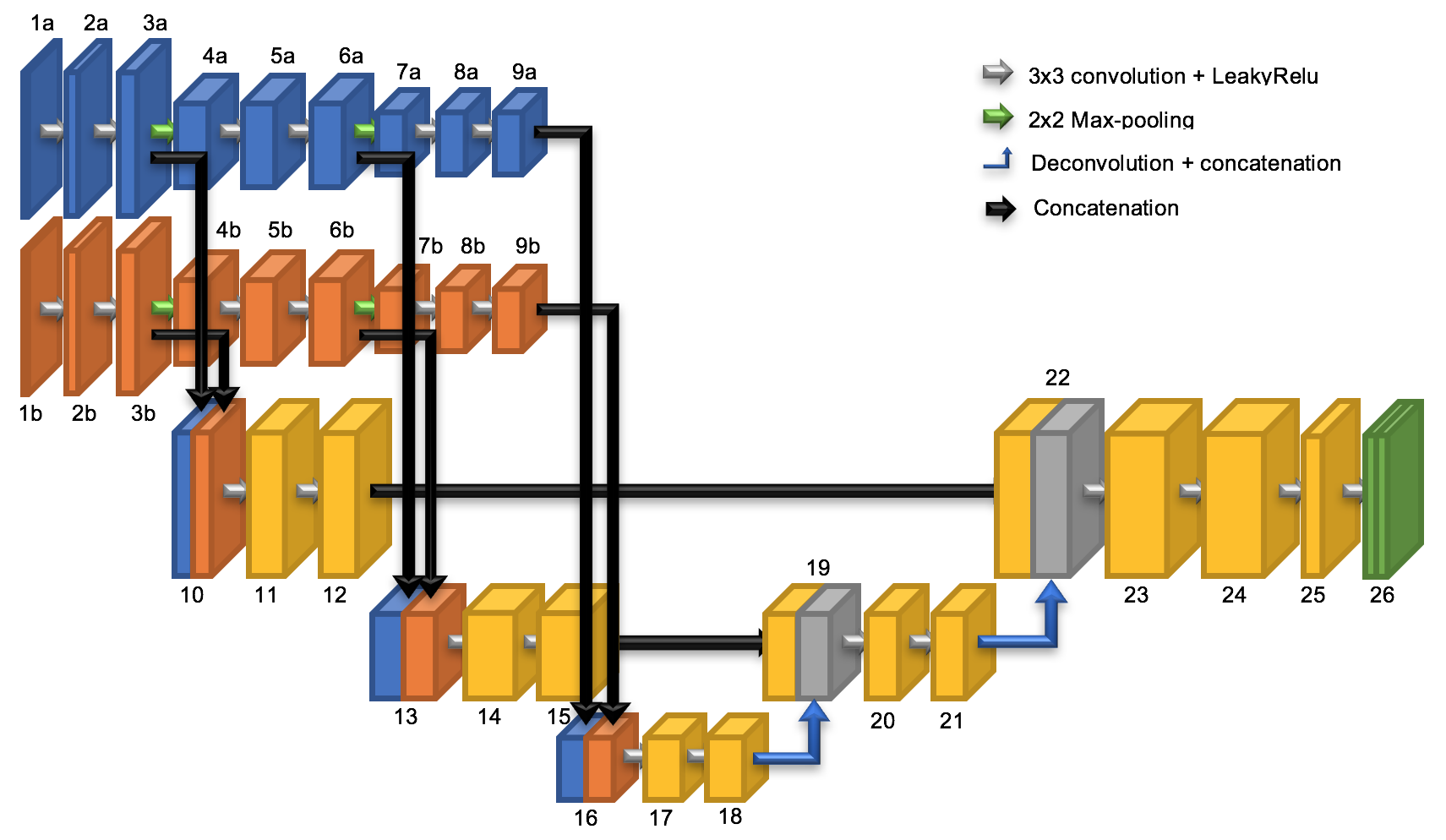}
\end{center}
   \caption{\label{fig:net}\textbf{Network architecture.} The two input images $I_1$ and $I_2$ are fed to layers 1a and 1b respectively. The output is a 2 dimensional vector map (layer 26 with 2 channels). See Appendix for all details.}
\end{figure}

\begin{figure*}[t]
\begin{center}
\begin{minipage}[l]{0.65\linewidth}
  \includegraphics[width=0.99\linewidth]{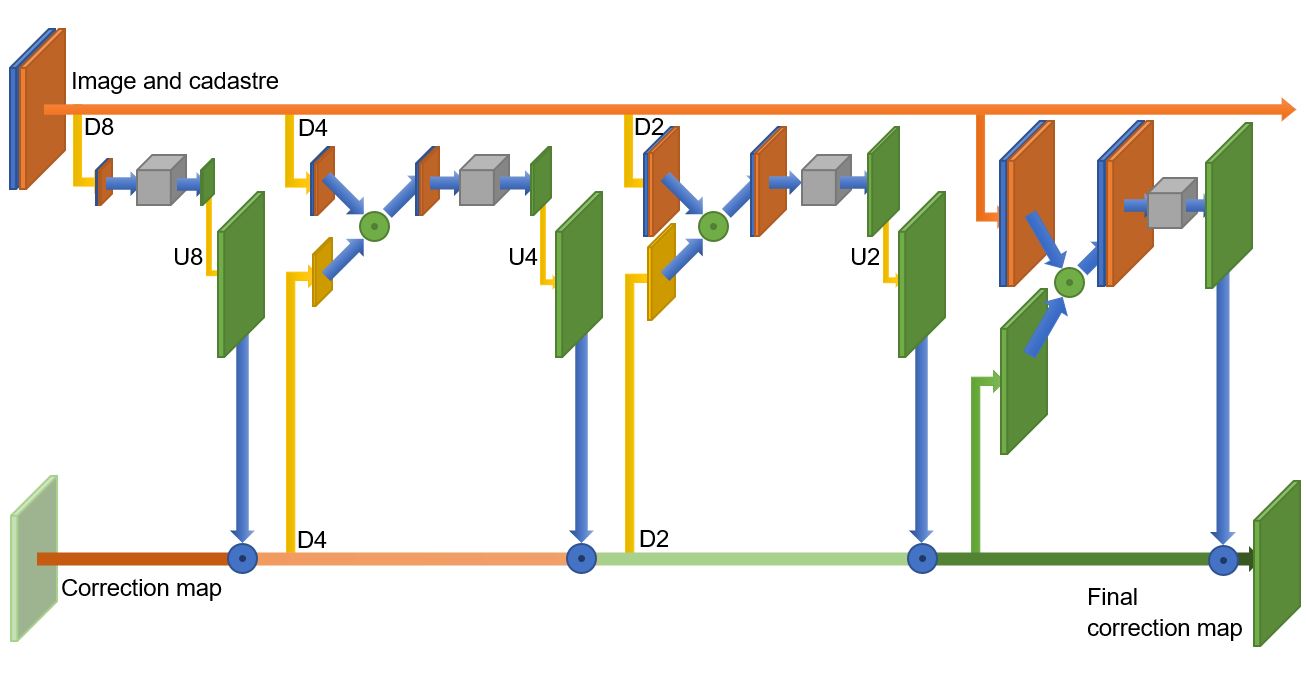}
  \end{minipage}
  \hfill
\begin{minipage}[r]{0.34\linewidth}
  \includegraphics[width=0.99\linewidth]{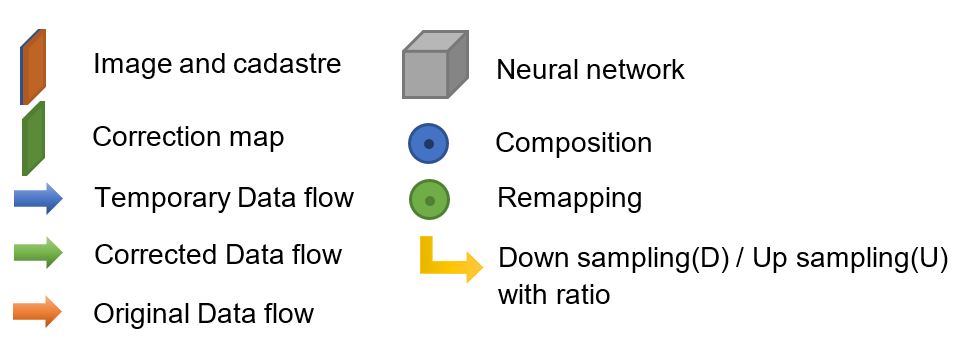}
  \end{minipage}
\end{center}
\caption{\label{fig:chain}\textbf{Full architecture as a chain of scale-specific neural networks.} The two full-resolution input images are always available on the top horizontal row. The full-resolution deformation is iteratively refined, scale per scale, on the bottom horizontal line. Each scale-specific block downsample the images to the right size, apply the previously-estimated deformation, and refines it, in a way somehow similar to ResNet. Each block is trained independently.}
\end{figure*}



\begin{figure}[h]
\begin{center}
   \includegraphics[width=0.49\linewidth]{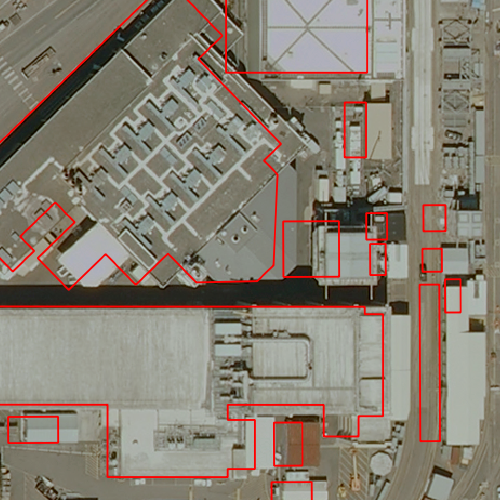}
   \includegraphics[width=0.49\linewidth]{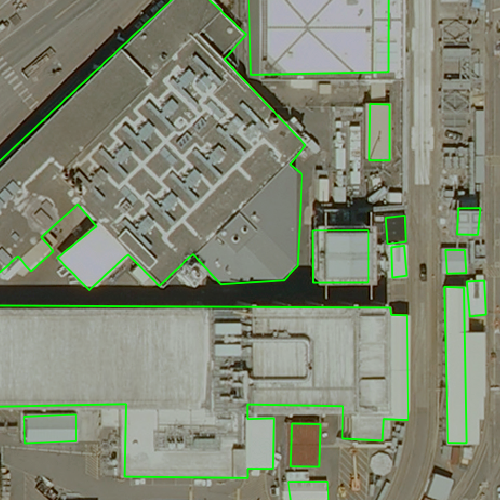}
   \caption{\label{fig:imageYT}\textbf{Example of image alignment.} Left : original image and OpenStreetMap (OSM) map. Right : Alignment result.}
\end{center}
\end{figure}

We perform four experiments on different datasets. 
The \textbf{first experiment} uses the \textbf{Inria aerial image labeling dataset}~\cite{maggiori2017dataset}, which is a collection of aerial orthorectified color (RGB) imagery with a spatial resolution of 30 cm/pixel covering 810 km$^2$ over 9 cities in USA and Austria. We aim at aligning a map of buildings downloaded from OSM with the images from the Inria dataset. The network described in Section \ref{sec:fullnet} is trained using image patches from six different cities, for which accurate building cadastral data are available\footnote{The cadastral data are extracted from OSM and contain a small misalignment of an order of several pixels.}. We then evaluated the network by using images of the area of Kitsap County not presented during training. Fig.~\ref{fig:imageYT} shows an example close-up of alignment result.

In the \textbf{second experiment}, the network trained in the first experiment is used to align the OSM building map with satellite images with a pansharpened resolution of 50 cm/pixel, acquired by the Pl\'eiades sensor over the Forez rural area in France.

To measure performance of the network, we use the percentage of correct key point metric~\cite{ign}. We manually identified matching key points on two couples of multimodal images (one Kitsap image from experiment 1 and one Forez image from experiment 2) with more than 600 keypoints for each image. We then measure the distance between the positions of keypoints after alignment by using different algorithms and the manually indicated ones. If this distance is smaller than a certain threshold, the keypoint is identified as matched.

\begin{figure}[h]
\begin{center}
  \begin{minipage}[l]{0.39\linewidth}
    \medskip
    
  $\!\!\!\!$\includegraphics[width=1.13\linewidth]{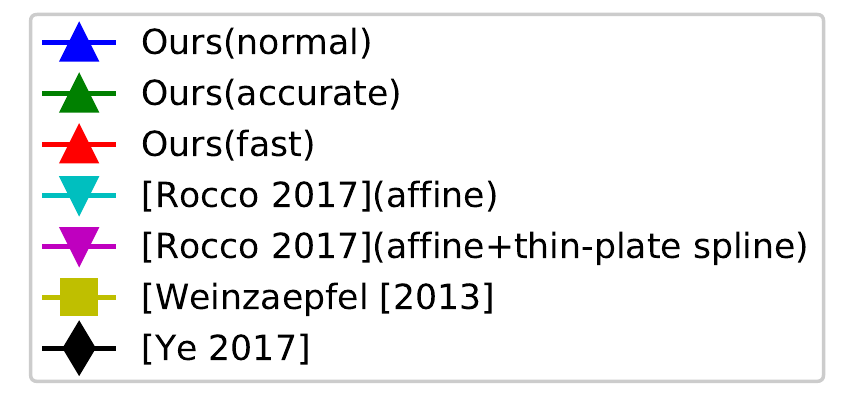}$\!\!\!\!$
  
  \vspace{1.5cm}
  
  \caption{\label{fig:longYT}\textbf{Key points matching.} Scores of different methods on the Kitsap and Forez datasets. The curves indicate the percentage of keypoints whose distance to the ground truth is less than the threshold in abscissa. Higher is better.}
  \end{minipage}
\begin{minipage}[r]{0.6\linewidth}
   \includegraphics[width=\linewidth]{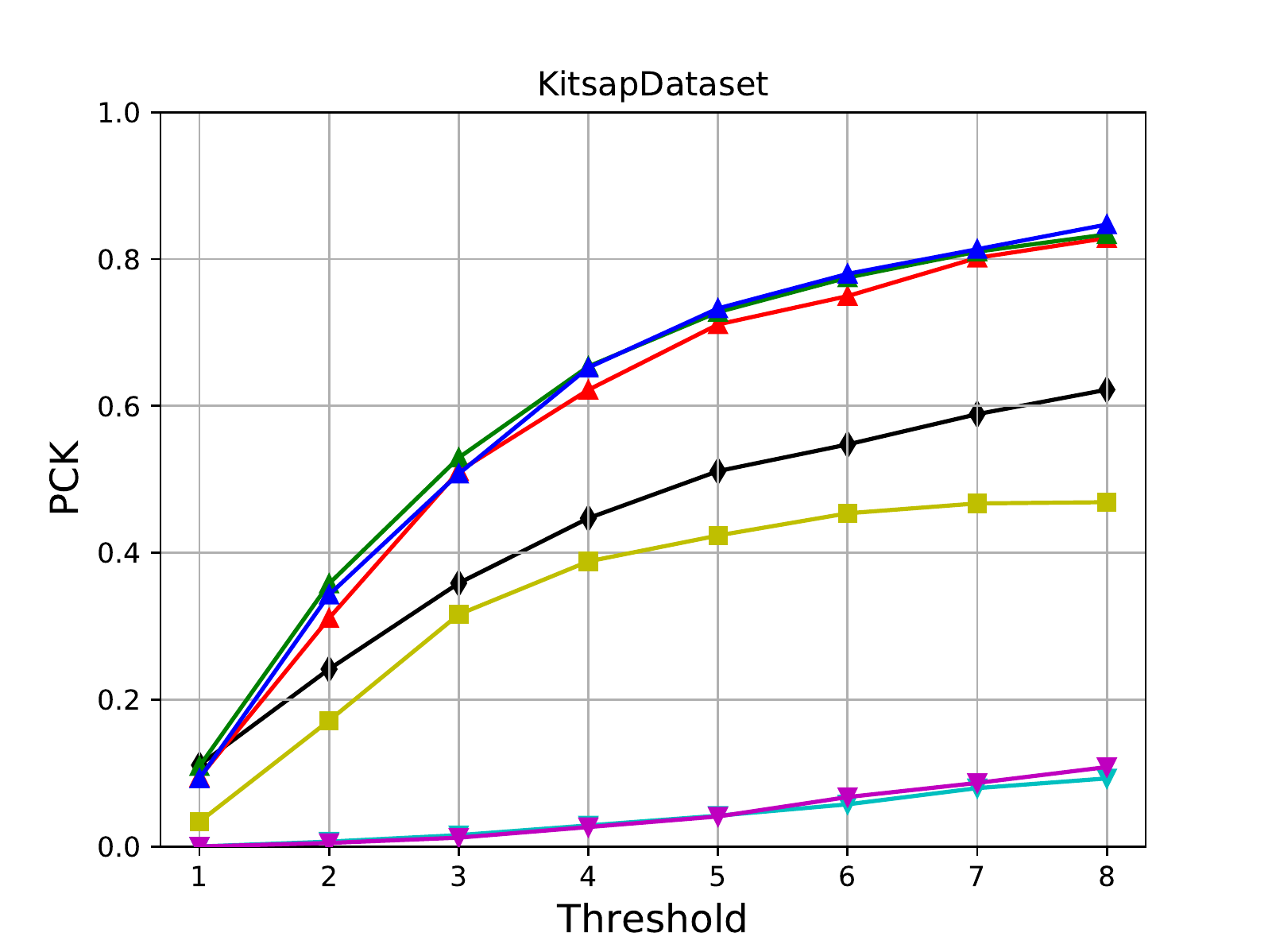}
   \includegraphics[width=\linewidth]{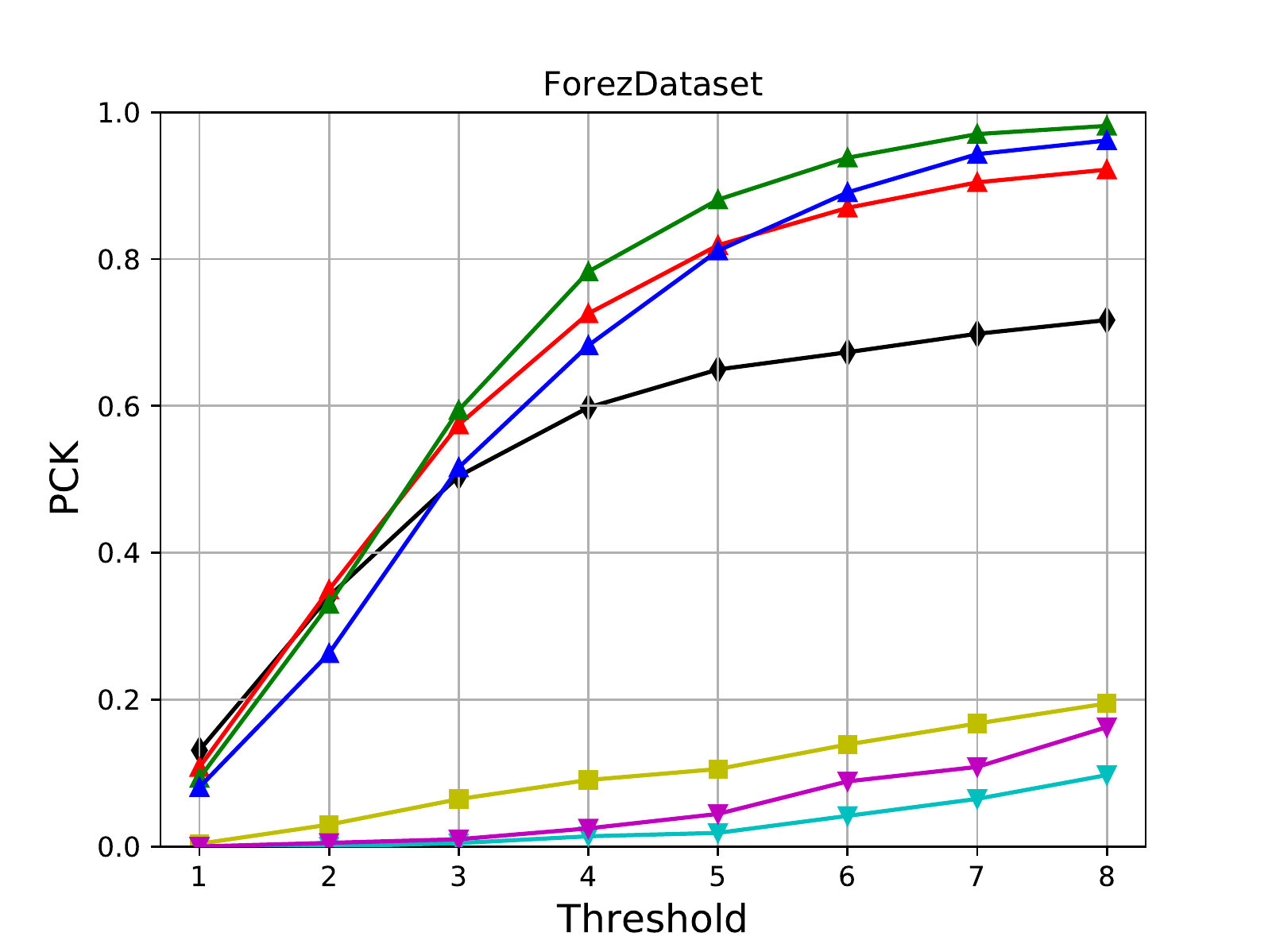}
 \end{minipage}
\end{center}
\end{figure}

\begin{figure}[h]
\begin{center}
   \begin{tabular}{@{}c@{}}
     \includegraphics[width=0.245\linewidth]{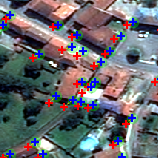}\\
     \includegraphics[width=0.245\linewidth]{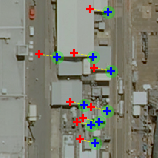}\\
   \scalebox{0.65}{(a) Ground truth}
   \end{tabular}
   \begin{tabular}{@{}c@{}}
     \includegraphics[width=0.245\linewidth]{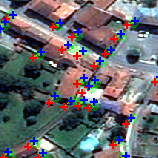}\\
     \includegraphics[width=0.245\linewidth]{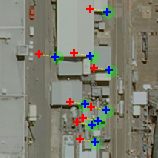}\\
   \scalebox{0.65}{(b) Ours}
   \end{tabular}
   \begin{tabular}{@{}c@{}}
     \includegraphics[width=0.245\linewidth]{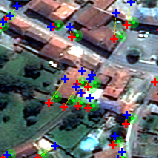}\\
     \includegraphics[width=0.245\linewidth]{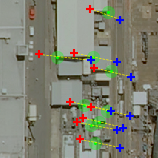}\\
     \scalebox{0.65}{(c) Rocco \cite{ign}}
   \end{tabular}
   \begin{tabular}{@{}c@{}}
     \includegraphics[width=0.245\linewidth]{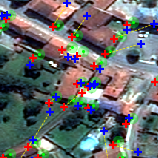}\\
     \includegraphics[width=0.245\linewidth]{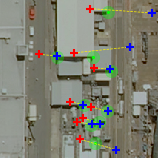}\\
   \scalebox{0.65}{(d)  Weinzaepfel \cite{deepMatching}} 
   \end{tabular}$\!\!$
   \caption{\label{fig:visuForez}\textbf{Multimodal keypoint matching comparison} for different methods and two datasets. Top: Forez dataset; bottom: Kitsap. Blue: predicted, green: ground truth. Full resolution in Appendix.
}
\end{center}
\end{figure}

Fig.~\ref{fig:longYT} compares the performance of our network with the following methods: DeepFlow of Weinzaepfel \textit{et al.}~\cite{deepMatching}, two variations of geometric matching method of Rocco et al.~\cite{ign}, and a multimodal registration method of Ye et al.~\cite{ye2017robust}. Our approach clearly outperforms other ones by a large margin. We note that averaging over rotations and symmetries (in green, ``\emph{accurate}'') does help on the Forez dataset, and that learning scale-specific features performs slightly better than scale-independent features but not always (blue vs.~red, ``\emph{fast}''). Examples of alignment results are shown in Figure \ref{fig:visuForez}. Our approach is also much faster, as shown by the computational times below for a $5000\times 5000$ image, even though we compute a dense registration while other approaches only match keypoints:\\
\noindent \begin{tabular}{|l|c|c|c|c|}
\hline
Method & Ours & \cite{ign} & \cite{deepMatching} & \cite{ye2017robust} \\
\hline
Time & {\bf 80 s} & 238 s & 784 s & 9550 s \\
\hline
CPU & Opteron 2Ghz & \multicolumn{2}{c|}{Intel
2.7Ghz} & 
Int. 3.5Ghz \\
\hline
GPU & GTX 1080 Ti & \multicolumn{2}{c|}{Q.M2000M} & GT 960 M \\
\hline
\end{tabular}

\begin{figure}[t]
\begin{center}
   \includegraphics[width=0.39\linewidth]{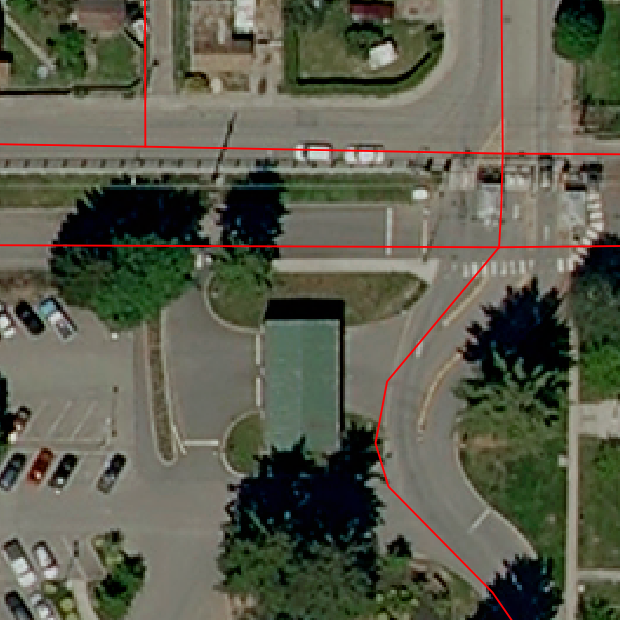}
   \includegraphics[width=0.39\linewidth]{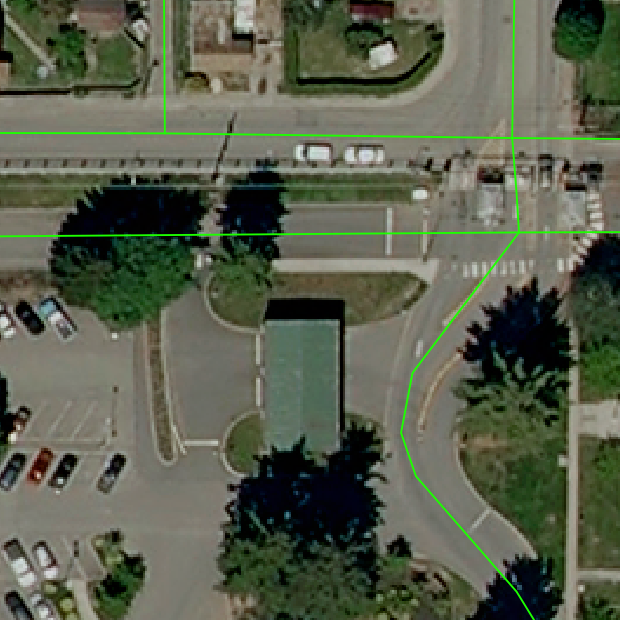}
   \includegraphics[width=0.39\linewidth]{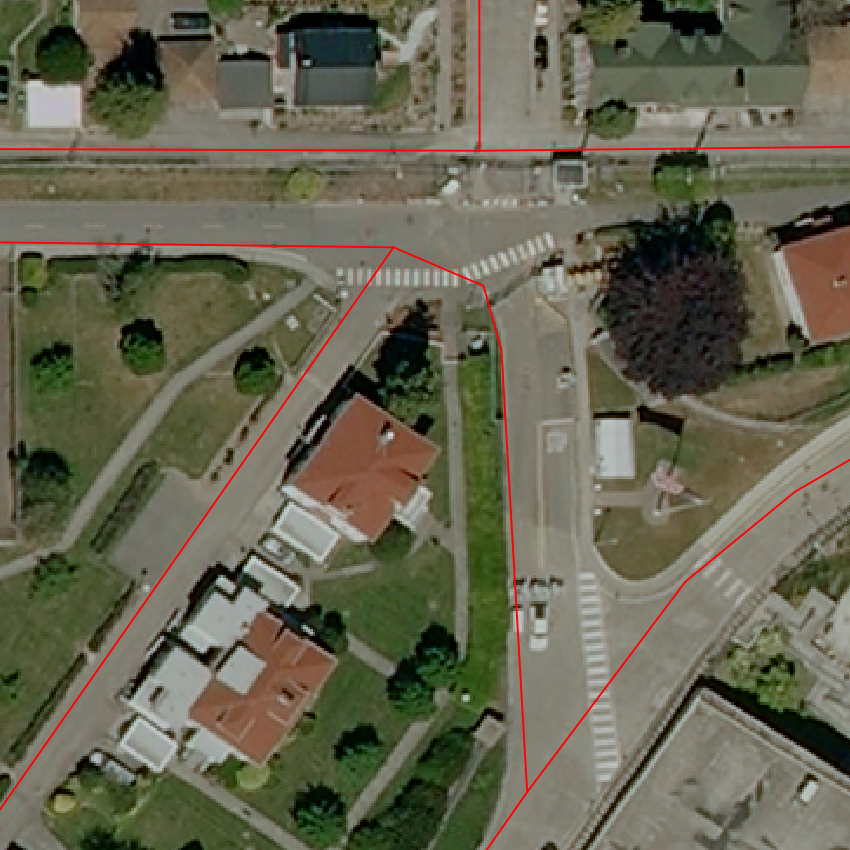}
   \includegraphics[width=0.39\linewidth]{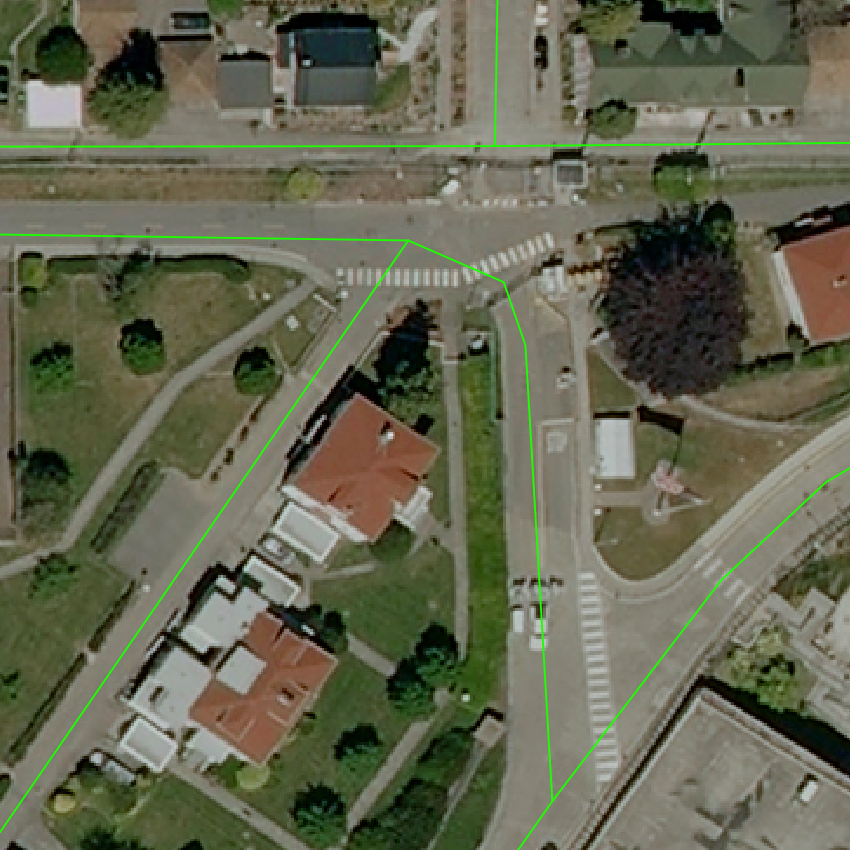}
   \caption{\label{fig:roadalign}\textbf{Example of road alignment.} Left: original alignment between image and roads (Kitsap); right: results after realignment.
}
\end{center}
\end{figure}

In a {\bf third experiment}, we align roads with the images used in the first experiment. The task differs from previous experiments in that only the center line of the road is known, and in the form of a polyline. Moreover, local edges are not useful features for alignment anymore, as the center of roads is homogeneous. We train on OSM data, by  dilating road polylines to reach a 4 pixel width and rasterizing them. We then test the performance of the trained network on the Kitsap image. The results are shown in Figure \ref{fig:roadalign}.

\begin{figure}[t]
\begin{center}
   \includegraphics[width=0.45\linewidth]{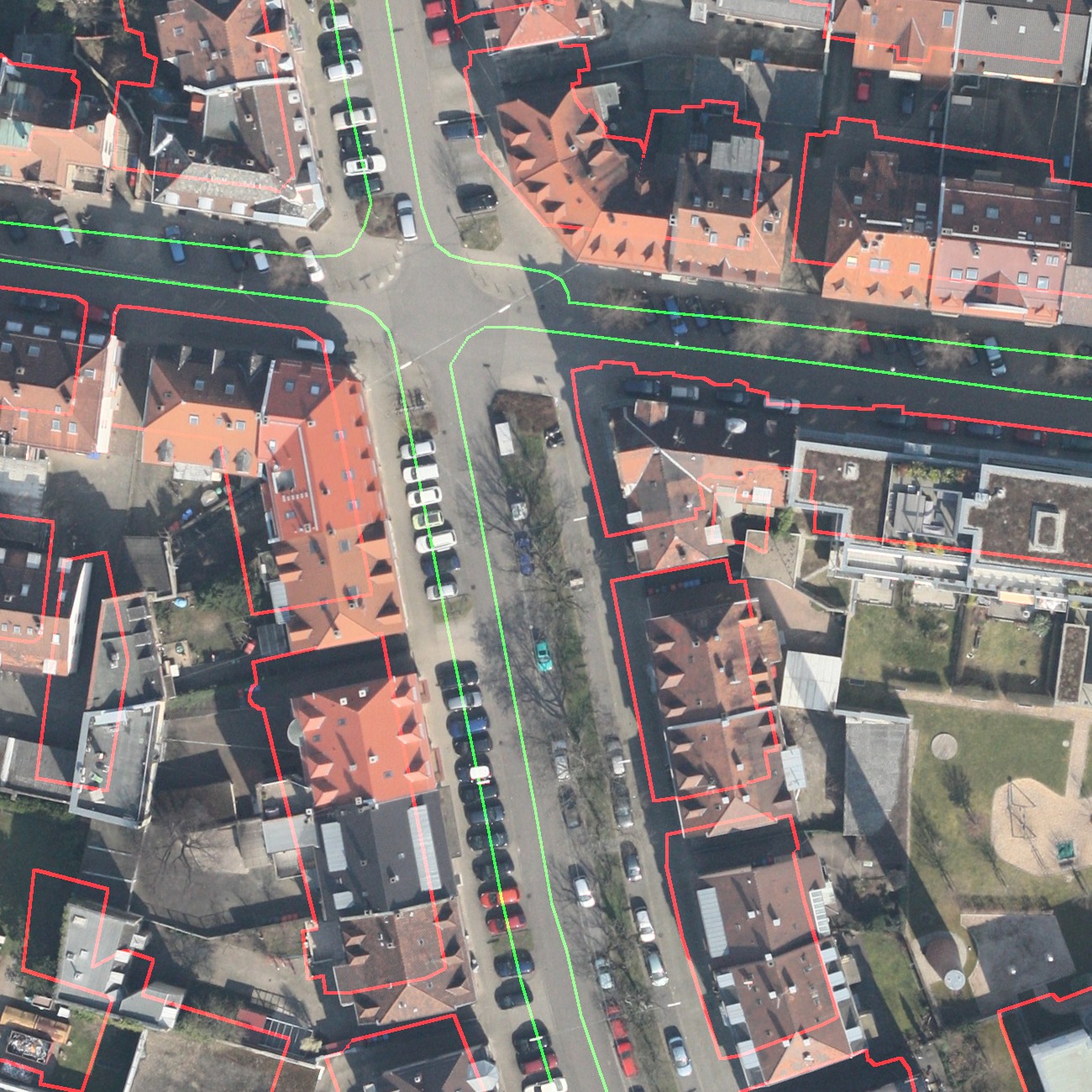}
   \includegraphics[width=0.45\linewidth]{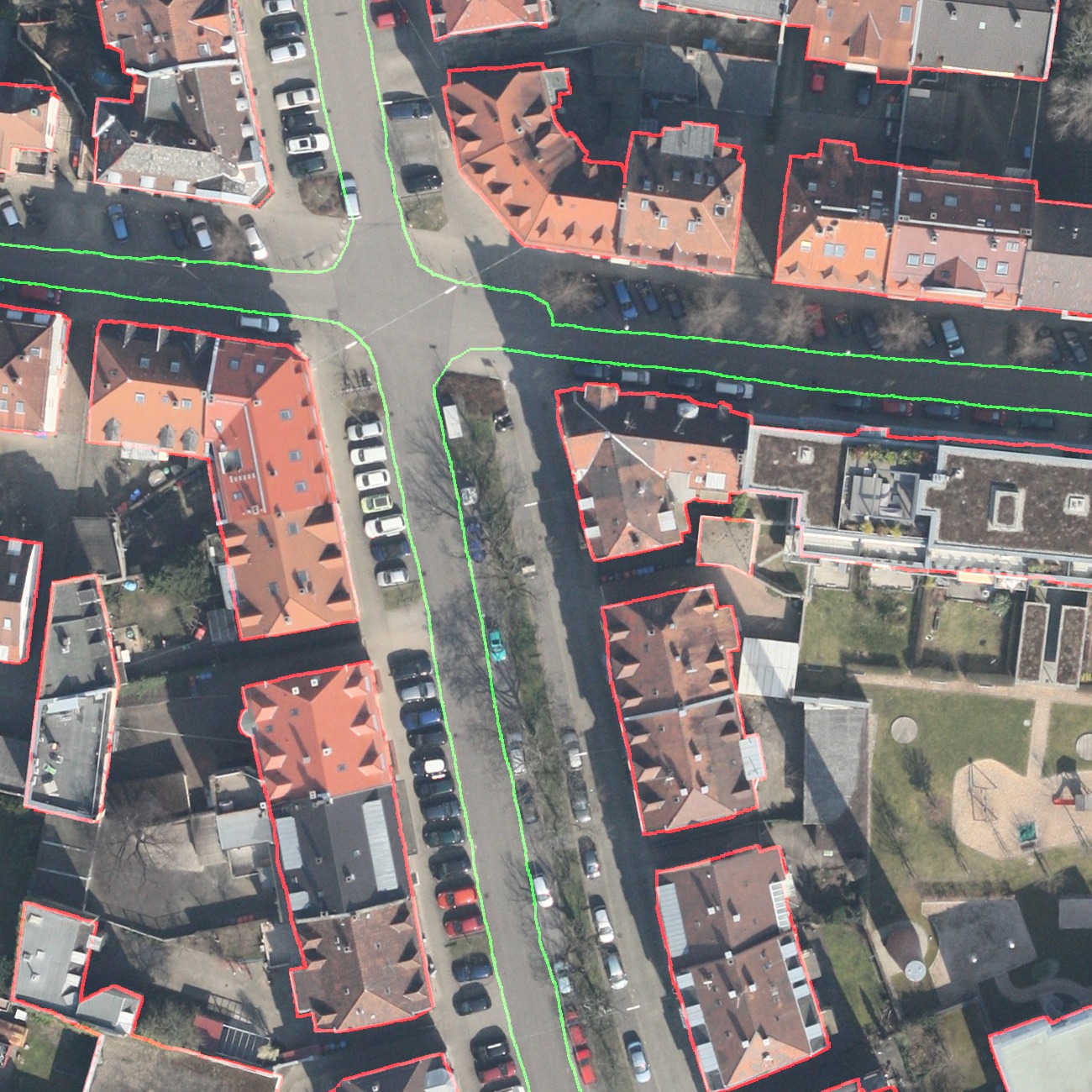}
   \caption{\label{fig:Kitty}\textbf{Example of alignment on the Kitti Dataset.} Left: before alignment; right: after alignment.
}
\end{center}
\end{figure}
\begin{figure}[h]
\begin{center}
$\!\!\!\!\!\!\!\!\!$   \includegraphics[width=0.56\linewidth]{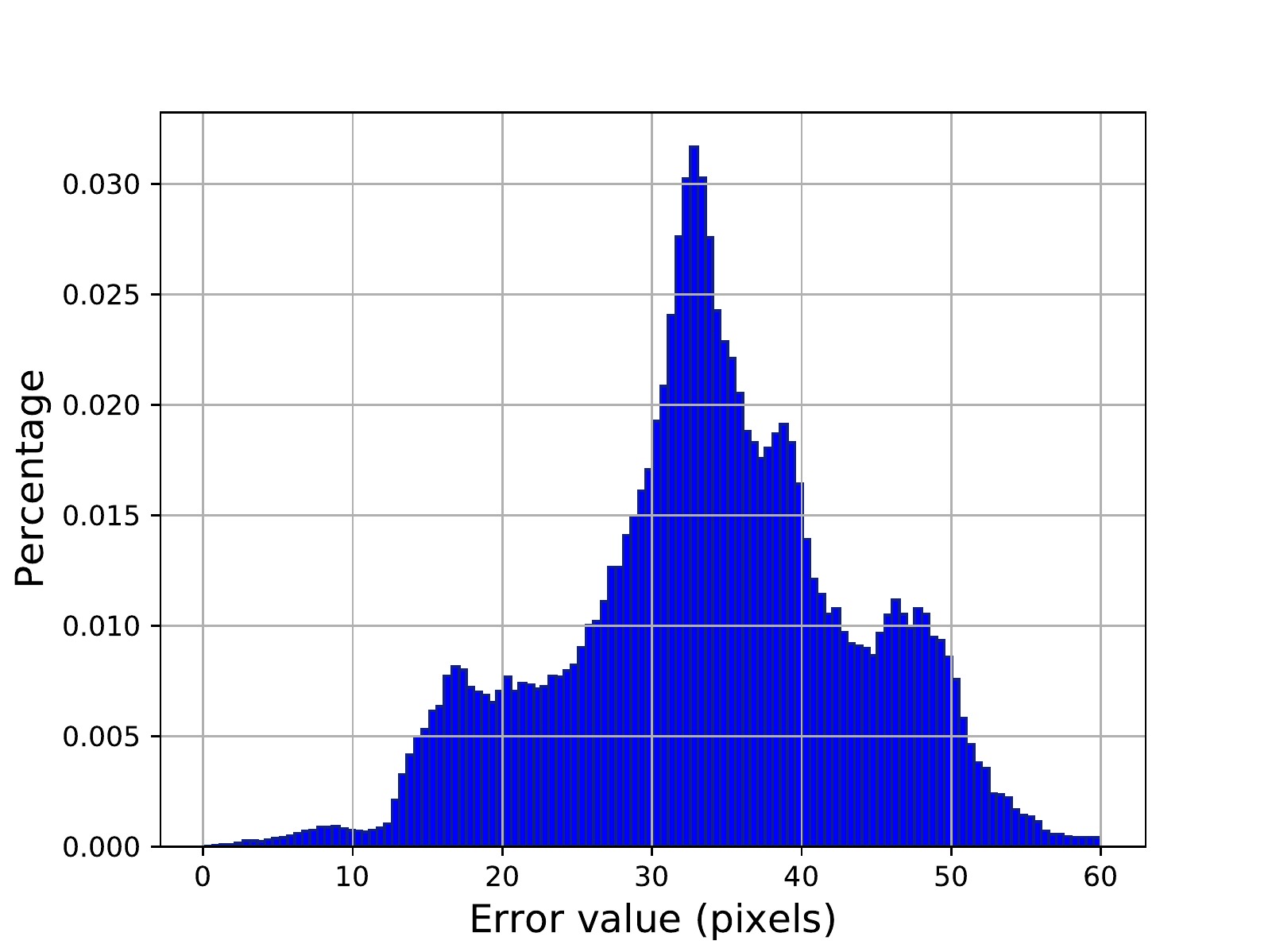} \hspace{-6mm}
   \includegraphics[width=0.56\linewidth]{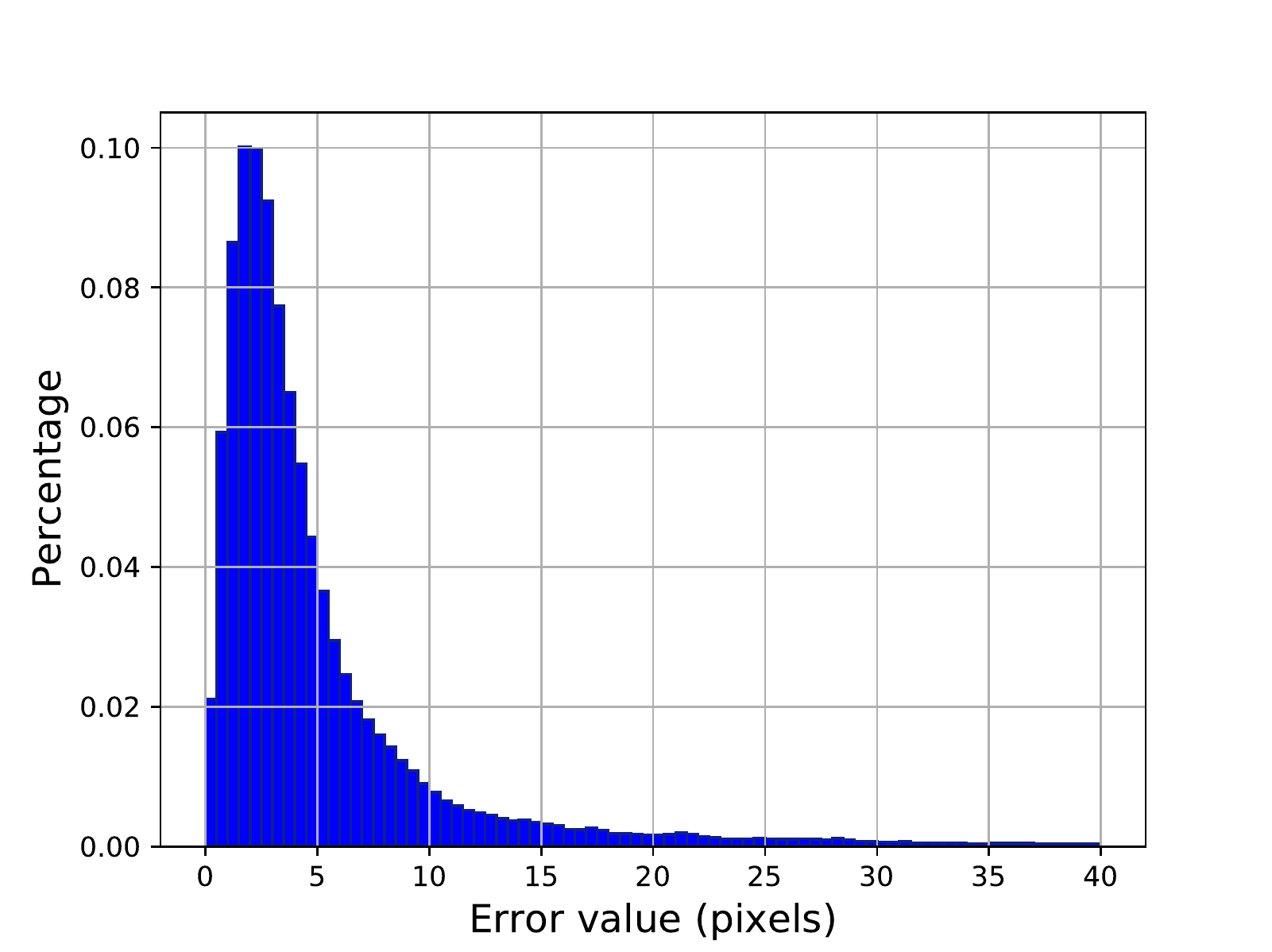}$\!\!\!\!\!\!\!\!\!\!\!\!$
   \caption{\label{fig:histo}\textbf{Misalignment histograms on the fourth experiment (Kitti dataset).} Left: original misalignment distribution in the test set; right: remaining error after our automatic alignment.
}
\end{center}
\end{figure}

The {\bf fourth experiment} checks the performance of our approach on a higher-resolution dataset. We consider the Kitti dataset \cite{geiger2013vision}, which contains high precision aerial images (9 cm/pixel), as well as perfectly aligned multi classes labeling \cite{mattyus2016hd}. We create a training set with artificial random deformations, in the same spirit as before, and a test set with randomly deformed images as well, but following different distributions, in order to check also the robustness of our training approach. Image pairs to be registered consist of a RGB image and a 3-channel binary image indicating buildings, roads and sidewalk presence respectively.
An example of result is shown in Figure \ref{fig:Kitty}. We also analyse the distribution of misalignments before and after registration, shown as histograms in Figure \ref{fig:histo}. We note that the vast majority of pixels are successfully very closely matched to their ground truth location.



We also perform an extra experiment to show that our multi-scale approach could generalize to other applications. We consider the problem of stereovision, where the input is a pair of RGB images taken from slightly different view points, and the expected output is the depth map, \ie a single channel image instead of a deformation field. We consider the dataset from \cite{stereoData1,stereoData2} and define the loss function as the depth error (squared), plus the regularizer (\ref{eq:laplacien}). We keep the same architecture but link the scale-specific networks with additions instead of compositions, so that each block adds scale-specific details to the depth map. The promising result (first run, no parameter tuning) is shown in the Appendix. 


\Black

\paragraph{Optimization details}
The network is trained with an Adam optimizer, on mini-batches of 16 patches of 128 $\times$ 128 pixels images,
with a learning rate starting from 0.001 and decayed by 4\% every 1000 iterations. Weights are initialized following Xavier Glorot's method.
We trained for 60 000 iterations.
More technical details are available in the Appendix. 

\paragraph{Additional details specific to sparse modalities} such as cadastral maps, though not essential.
During training, we sort out fully or mostly blank images (\eg cadastre without any building). Also, in order to train more where there is more information to extract (\eg, corners and edges vs.~wide homogeneous spaces), we multiply the pixel loss by a factor $>1$ on building edges when training.


When rectangular building are glued together in a row with shared walls, the location of their edges and corners is not visible anymore on the rasterized version of the OSM cadastre. By adding a channel to the cadastre map, reminding the OSM corner locations, we observe a better alignment of such rows.

%% file: discussion.tex
\section{Conclusion}

Based on an analysis of classical methods, we designed a chain of scale-specific neural networks for non-rigid image registration. By predicting directly the final registration at each scale, we avoid slow iterative processes such as gradient descent schemes. The computational complexity is linear in the image size, and far lower than even keypoint matching approaches. We demonstrated its performance on various remote sensing tasks and resolutions. The trained network as well as the training code will be made available online. This way, we hope to contribute to the creation of large datasets in remote sensing, where precision so far was an issue requiring hand-made ground truth.

An interesting point to study further is the specialization of the networks for each scale, to check on which type of image dataset it is strong or not (medical imaging vs.~landscape pictures \eg).



%% file: annexe.tex
\subsection{Neural network architecture details}

\begin{figure*}[t]
\begin{center}
   \includegraphics[width=0.75\linewidth]{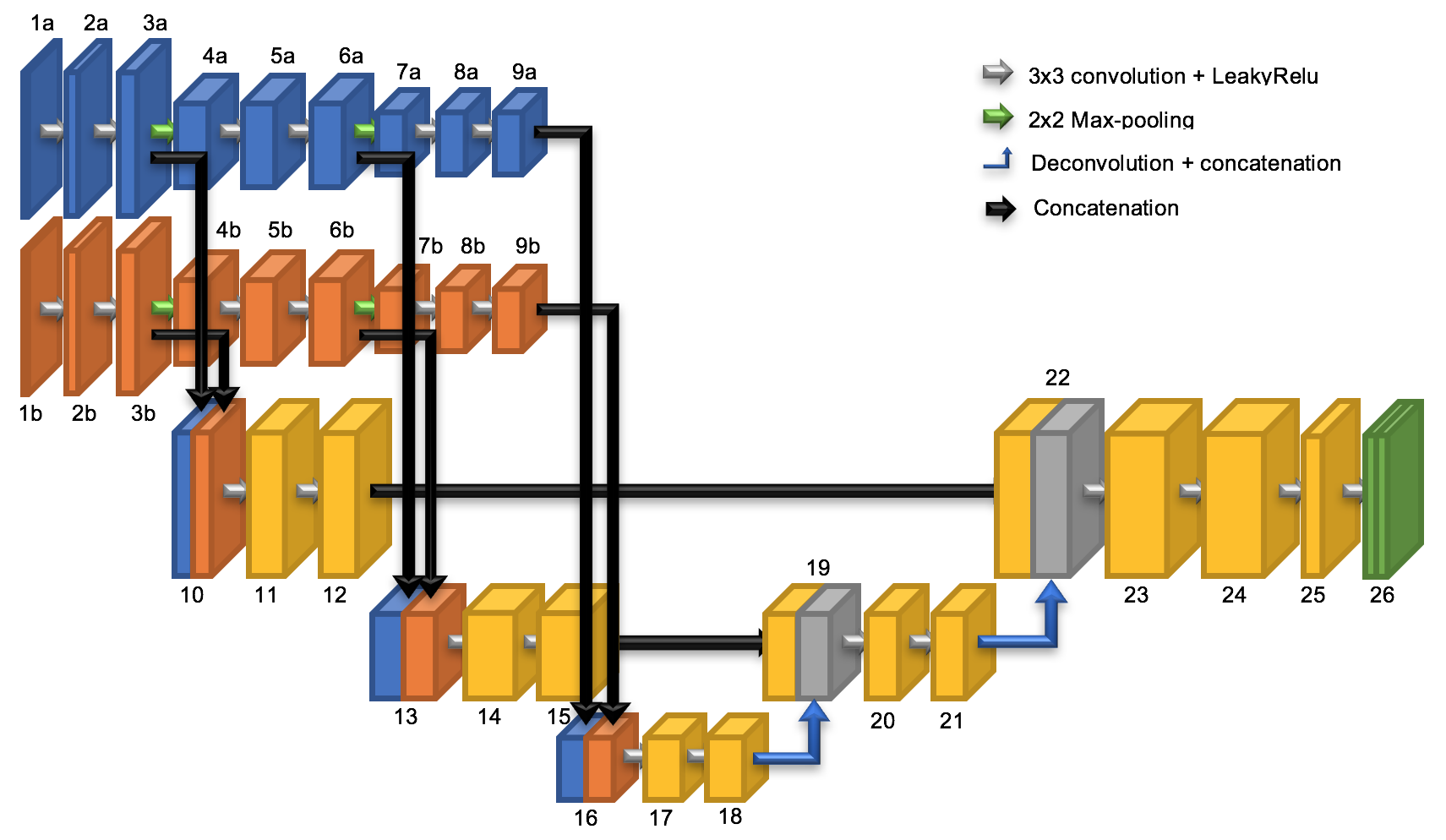}
\end{center}
   \caption{\label{fig:net2} \textbf{Network architecture.}
The two input images $I_1$ and $I_2$ are fed to layers 1a and 1b respectively. The output is a 2 dimensional vector map (layer 26 with 2 channels). This architecture allows to merge information from both sources at all scales, to extract high-level information, and to remember fine details from the input resolution to output a precise full-resolution deformation.
Details on Figure \ref{fig:NNdetail}.} \vspace{5mm}

\begin{tabular}{|c|c|l|c|c|c|c|l|}
  \hline
  Start layer & End Layer & Name & Kernel size & Number of filters & padding & stride \\
  \hline
  1 & 2 & convolution-1 & 5 & 16 & 2 &  \\
  2 & 3 & convolution-2 & 5 & 32 & 2 &  \\
  3 & 4 & pooling-1 & 2 &  &  & 2  \\
  4 & 5 & convolution-3 & 3 & 32 & 1 &  \\
  5 & 6 & convolution-4 & 3 & 32 & 1 &  \\
  6 & 7 & pooling-2 & 2 &  &  & 2  \\
  7 & 8 & convolution-5 & 3 & 64 & 1 &  \\
  8 & 9 & convolution-6 & 3 & 64 & 1 &  \\
  3 & 10 & concatenation-1 &  &  &  &   \\
  6 & 13 & concatenation-2 &  &  &  &   \\
  9 & 16 & concatenation-3 &  &  &  &   \\
  10 & 11 & convolution-7 & 3 & 32 & 1 &   \\
  11 & 12 & convolution-8 & 3 & 32 & 1 &   \\
  13 & 14 & convolution-9 & 3 & 64 & 1 &   \\
  14 & 15 & convolution-10 & 3 & 64 & 1 &   \\
  16 & 17 & convolution-11 & 3 & 64 & 1 &   \\
  18 & 19 & convolution-12 & 3 & 64 & 1 &   \\
  18 & 18' & deconvolution-1 & 3 & 64 &  &   \\
  15-18' & 19 & concatenation-4 &  &  &  &   \\
  19 & 20 & convolution-11 & 3 & 64 & 1 &   \\
  20 & 21 & convolution-12 & 3 & 64 & 1 &   \\
  21 & 21' & deconvolution-2 & 3 & 32 &  &   \\
  12-21' & 22 & concatenation-5 &  &  &  &   \\
  22 & 23 & convolution-13 & 3 & 64 & 1 &   \\
  23 & 24 & convolution-14 & 3 & 64 & 1 &   \\
  24 & 25 & convolution-15 & 3 & 32 & 1 &   \\
  25 & 26 & convolution-16 & 3 & 2 & 1 &   \\
  \hline
\end{tabular}\medskip
\caption{\label{fig:NNdetail} Details for each layer of the (scale-specific) neural network displayed on Figure \ref{fig:net}. ``Kernel size 3'' for a convolutional layer means ``$3 \times 3$'' convolution. \vspace{-5mm}}
\end{figure*}

See Figures \ref{fig:net2} and \ref{fig:NNdetail} for further details about the architecture and meta-parameters.

\subsection{Alignment framework}

\begin{figure*}[p]
\begin{center}
     \includegraphics[width=0.8\linewidth]{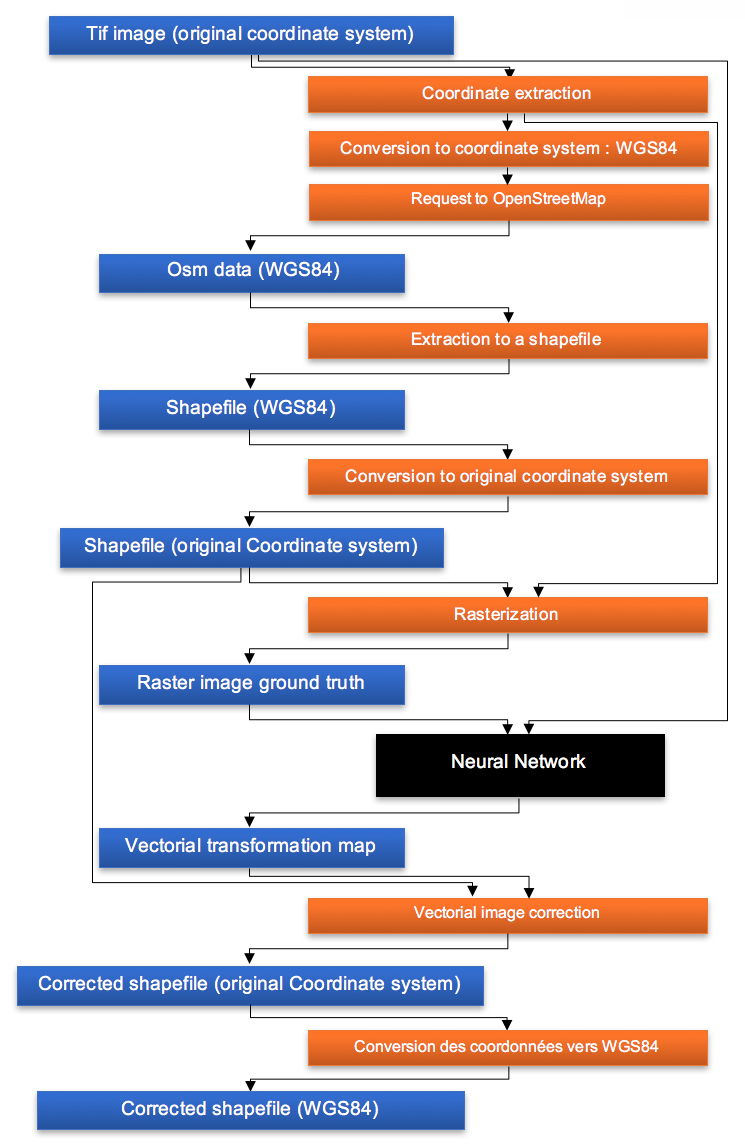}
     \caption{\label{fig:chart}\textbf{Global framework for OpenStreetMap data correction.}}
\end{center}
\end{figure*}

The whole processing framework for the alignment of OpenStreetMap cadastral information with aerial images is summarized in the chart shown in Figure 
\ref{fig:chart}.

\subsection{Example of deformation}

As explained in the article, to augment the dataset size for training, we generate random deformations, as a mixture of Gaussian functions with random shifts. An example of such a deformation (amplified 4 times for better visualisation purposes), and the result of its application to the original image, are shown in Figure \ref{fig:defo}.

\subsection{Training details}

\subsubsection{Tricks for better training}

To  reduce the training time and possible memory issues, patch of images ($256\times 256$ pixels) were given to the network for the training instead 
of the whole image, reducing the amount of computations needed per mini-batch. This is also important in terms of memory usage of the network as  
original images contain $5000\times 5000$ pixels. Furthermore, neural network computations and data generation (a 
random transformation is generated for each image at each training step in order to augment the training set size) are parallelized in  
order to improve the training speed of the algorithm. \\

Another issue encountered was to reach the local minimum  
corresponding to  outputting always a null deformation, thus preventing the neural network parameters to evolve towards a better optimum.  
To solve this issue we used several methods to facilitate the training of the network and reduce its probability to reach  
this local minima.\\
The first technique used was to force the network to overfit a very small sample of the dataset (400 iterations on 
4 images with random transformations). This proved to 
be particularly efficient at avoiding the null local minimum.

The second technique is specific to the dataset used, \ie the data from the cadastral images is particularly sparse. The first step is then to 
check, before selecting any training example, that data needed for alignment is present on the cadastral image, \ie the cadastral image does contain buildings, \eg. This was done simply by 
calculating the ratio between labeled and unlabeled pixels (cadastral/non-cadastral) on each candidate image patch, and by setting a range of 
accepted values (\eg, the minor class of an image should represent at least 5\% of the labels of this image). Patches
not respecting this rule were not selected when forming mini-batches.
Thus mini-batches contained only relevant examples.

\begin{figure}[!hb]
\begin{center}
   \includegraphics[width=0.45\linewidth]{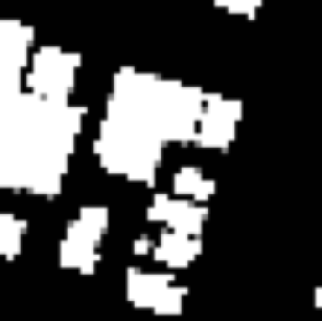}
   \includegraphics[width=0.45\linewidth]{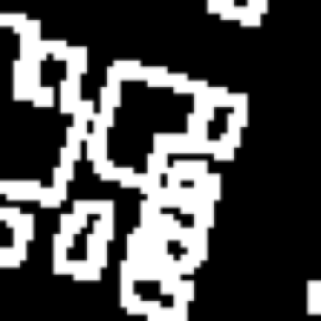}
   \includegraphics[width=0.45\linewidth]{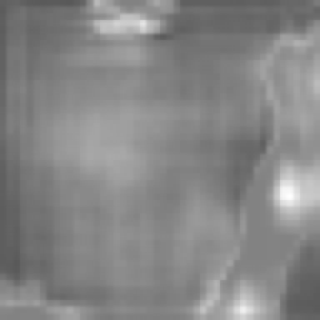}
   \includegraphics[width=0.45\linewidth]{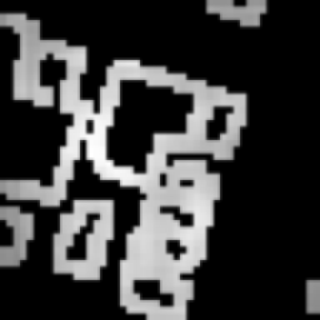}
   \caption{\label{fig:mark}\textbf{Building boundaries used to re-weight the loss function.}  Top left: cadastral image; top right: building boundary mask used; bottom left: the original map of the loss function; bottom right: masked map of the loss function on boundaries. The loss will be multiplied with a constant factor in these areas.}
\end{center}
\end{figure}

\begin{figure*}[h]
\begin{center}
   \includegraphics[width=0.25\linewidth]{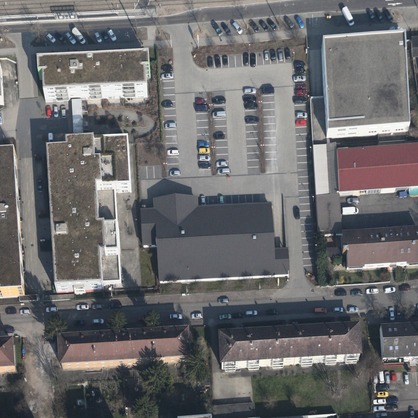}
   \raisebox{1.17mm}[0pt][0pt]{\fbox{\includegraphics[trim={21mm 5mm 2cm 5mm},clip,width=0.24\linewidth]{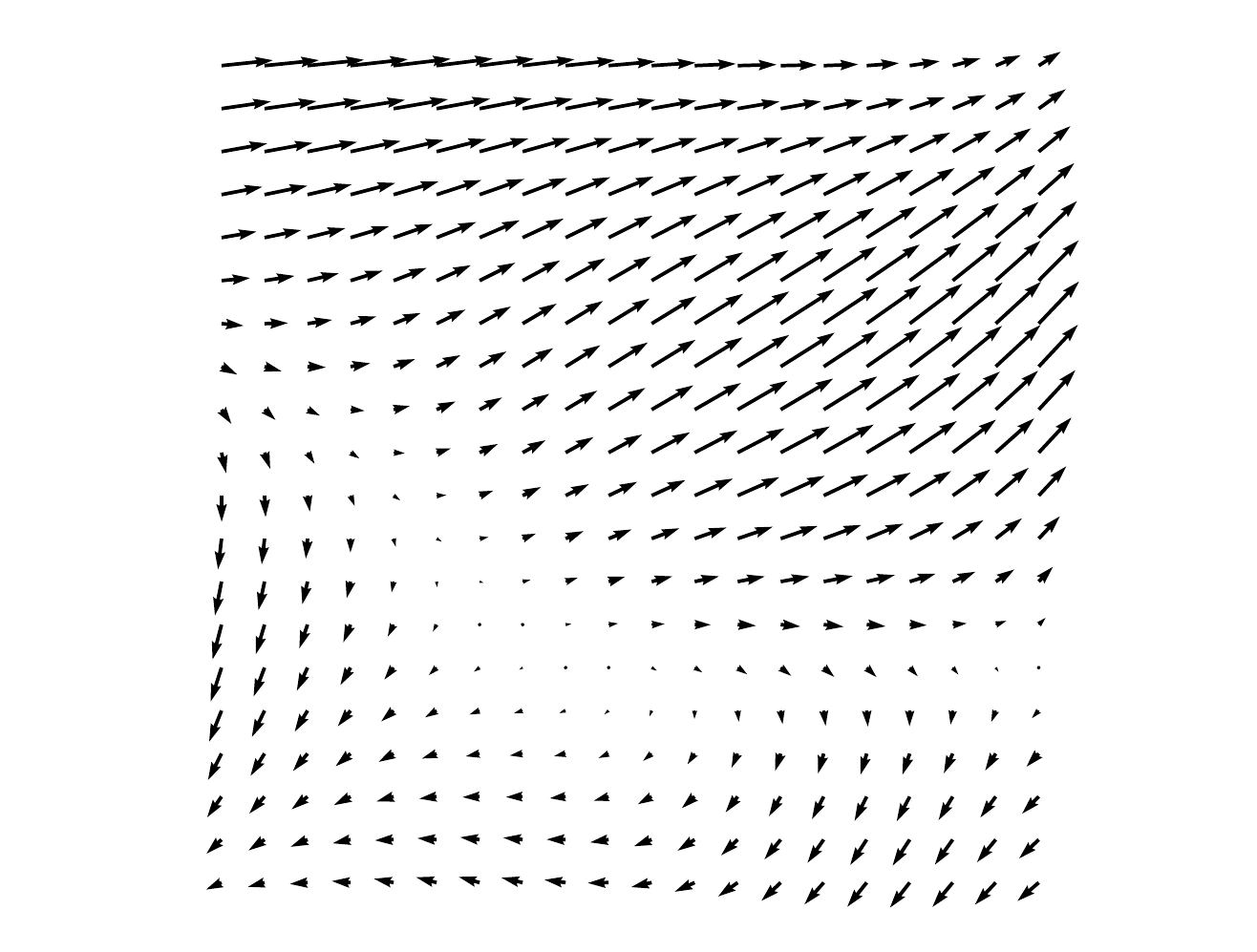}}}
   \includegraphics[width=0.25\linewidth]{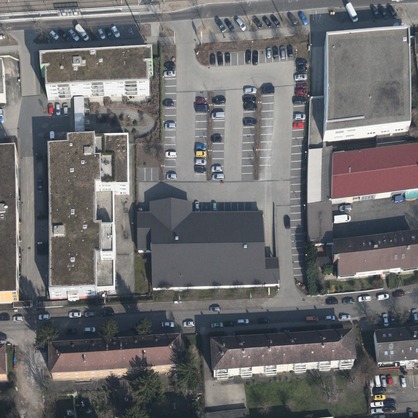}
\end{center}
\caption{\label{fig:defo}\textbf{Example of deformation.} Left: an image $I$; middle: a deformation $\phi$, \ie a $\R^2$ vector field; right: the associated deformed image $I\circ\phi$.}\vspace{2cm}
\end{figure*}

Another issue linked to the sparsity of the 
cadastre arises when the network is training on parts of a patch where not enough data is present to determine the 
transformation needed (\eg, only one class within an area, as in a garden for example), even to a human eye:
the estimation of the offset was not possible due to the lack of information. 
To solve this problem, we increase the weight of the loss 
function on the boundaries of buildings (parts where the transformation can be well estimated). For this, we first detect building boundaries based on the cadastre, as shown in \ref{fig:mark}, and add an multiplying factor to the loss at such locations (which is equivalent to sampling more often there).
This insures that the deformation is findable 
and that the training is useful. This last trick is specific to our dataset, which is binary labelled, but we show in experiments that 
this is not needed when the dataset is not sparse or binary. 

Lastly, we observed minor aligning issues when dealing with rows of houses with common walls, due to local translation invariance of the images, which adds locally a degree of freedom for alignment along the row axis.
We decided to give supplementary information corresponding to the location of all corners extracted from the OpenStreetMap vectorial image (as each corner of each house is indicated), hoping to help to guide the alignment along such translation-invariant line in the cadastre.
This step is however not critical as the results improvement is small and specific to certain building geometries (row of identical houses with common walls).

\subsubsection{Training information}

\noindent Number of iterations :          60 000 \\
Batch size :                   16 \\
Time to train :                16 hours \\
Number of images :             108 original images (with a random transformation generated at each iteration for each image) \\
Original image size :          $5000\times 5000$ \\
Patch size :                   $256\times 256$ \\
Total number of layers :       26 \\
Memory used with tensorflow :  9.7 GB \\
GPU :                          GeForce GTX 1080 \\
Processor :                    Dual-Xeon E5-2630 \\
RAM  :                         64 GB

\begin{figure*}[b]
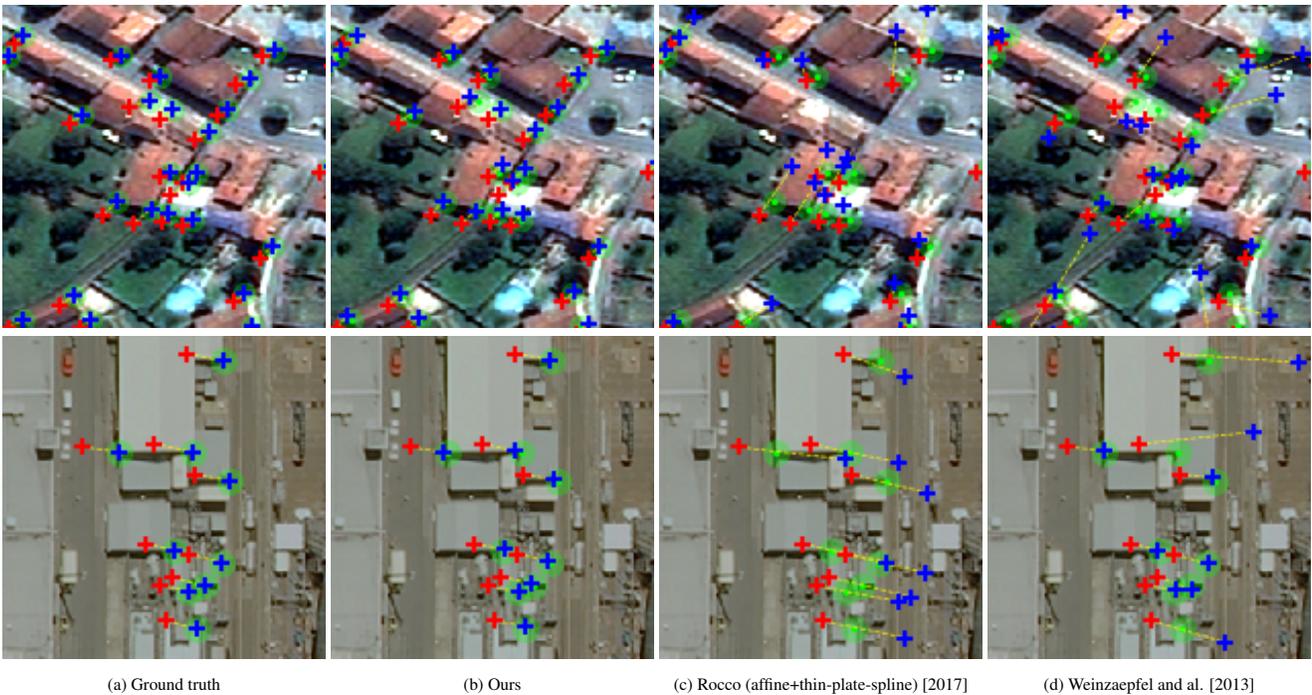

\begin{center}
   \begin{tabular}{@{}c@{}}
     \includegraphics[width=0.245\linewidth]{BTPsep_15GT.png}\\
     \includegraphics[width=0.245\linewidth]{kitsap25_25GT.png}\\
   \scalebox{0.65}{(a) Ground truth}
   \end{tabular}
   \begin{tabular}{@{}c@{}}
     \includegraphics[width=0.245\linewidth]{BTPsep_15PA.png}\\
     \includegraphics[width=0.245\linewidth]{kitsap25_25PA.png}\\
   \scalebox{0.65}{(b) Ours}
   \end{tabular}
   \begin{tabular}{@{}c@{}}
     \includegraphics[width=0.245\linewidth]{BTPsep_15i2.png}\\
     \includegraphics[width=0.245\linewidth]{kitsap25_25i2.png}\\
   \scalebox{0.65}{(c) Rocco (affine+thin-plate-spline) [2017]}
   \end{tabular}
   \begin{tabular}{@{}c@{}}
     \includegraphics[width=0.245\linewidth]{BTPsep_15D.png}\\
     \includegraphics[width=0.245\linewidth]{kitsap25_25D.png}\\
   \scalebox{0.65}{(d)  Weinzaepfel and al. [2013]}
   \end{tabular}
   \caption{\label{fig:keypointmatch}\textbf{Multimodal keypoint matching comparison} for different methods and two datasets. Top: Forez dataset; bottom: Kitsap. Blue: predicted, green: ground truth (centers of the green circles), red: original location of the corner (from the OpenStreetMap cadastral image, which is mis-geolocalized with respect to the RGB image).}
\end{center}
\end{figure*}

\subsection{Keypoint matching}

The keypoint matching experiment shown in the figure 7 of the main paper
is shown full resolution here in Figure \ref{fig:keypointmatch}.\\

\begin{figure*}[t]
\begin{center}
   \begin{tabular}{@{}c@{}}
     \rotatebox{90}{\tiny Our method}
     \includegraphics[width=0.2\linewidth]{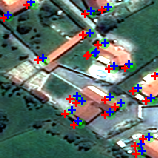}\\
     \rotatebox{90}{\tiny Our method (fast)}
     \includegraphics[width=0.2\linewidth]{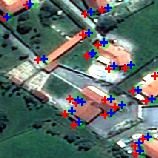}\\
     \rotatebox{90}{\tiny Rocco [2017] (affine)}
     \includegraphics[width=0.2\linewidth]{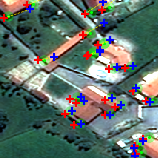}\\
     \rotatebox{90}{\tiny Rocco [2017] (affine+thin-plate spline)}
     \includegraphics[width=0.2\linewidth]{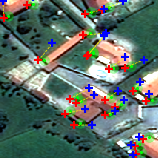}\\
     \rotatebox{90}{\tiny Weinzaepfel and al. [2013]}
     \includegraphics[width=0.2\linewidth]{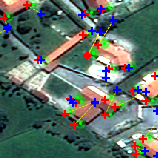}\\
     \rotatebox{90}{\tiny Ye [2017]}
     \includegraphics[width=0.2\linewidth]{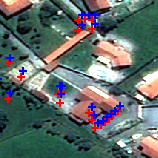}\\
   \end{tabular}
   \begin{tabular}{@{}c@{}}
     \includegraphics[width=0.2\linewidth]{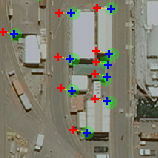}\\
     \includegraphics[width=0.2\linewidth]{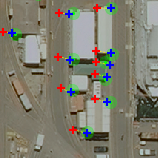}\\
     \includegraphics[width=0.2\linewidth]{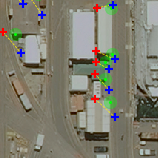}\\
     \includegraphics[width=0.2\linewidth]{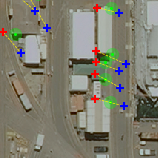}\\
     \includegraphics[width=0.2\linewidth]{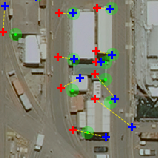}\\
     \includegraphics[width=0.2\linewidth]{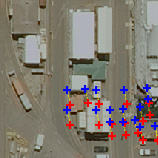}\\
   \end{tabular}
   \begin{tabular}{@{}c@{}}
     \includegraphics[width=0.2\linewidth]{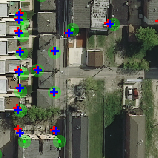}\\
     \includegraphics[width=0.2\linewidth]{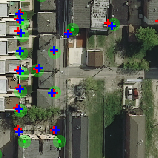}\\
     \includegraphics[width=0.2\linewidth]{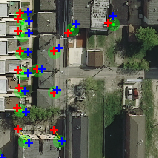}\\
     \includegraphics[width=0.2\linewidth]{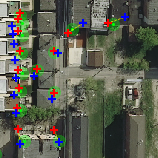}\\
     \includegraphics[width=0.2\linewidth]{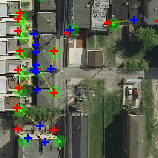}\\
     \includegraphics[width=0.2\linewidth]{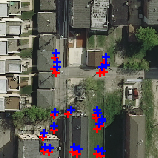}\\
   \end{tabular}
   \begin{tabular}{@{}c@{}}
     \includegraphics[width=0.2\linewidth]{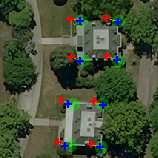}\\
     \includegraphics[width=0.2\linewidth]{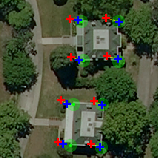}\\
     \includegraphics[width=0.2\linewidth]{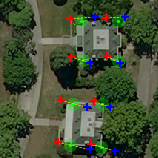}\\
     \includegraphics[width=0.2\linewidth]{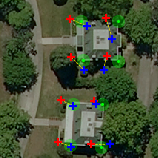}\\
     \includegraphics[width=0.2\linewidth]{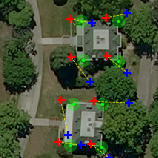}\\
     \includegraphics[width=0.2\linewidth]{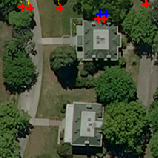}\\
   \end{tabular}
   \caption{\label{fig:suppmatch}\textbf{Additional multimodal keypoint matching examples.} Same setup as in Figure \ref{fig:keypointmatch}.}
\end{center}
\end{figure*}

Supplementary matching examples are shown in Figure~\ref{fig:suppmatch}.

\begin{figure*}[b]
\begin{center}
   \includegraphics[width=0.95\linewidth]{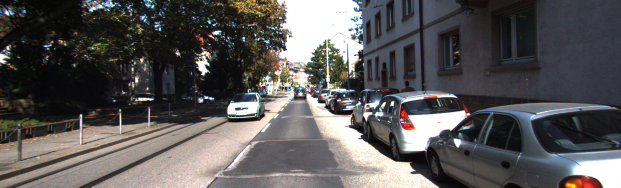}
   \includegraphics[width=0.95\linewidth]{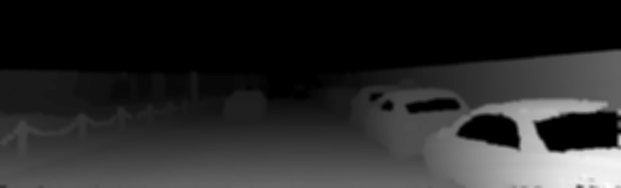}
   \includegraphics[width=0.95\linewidth]{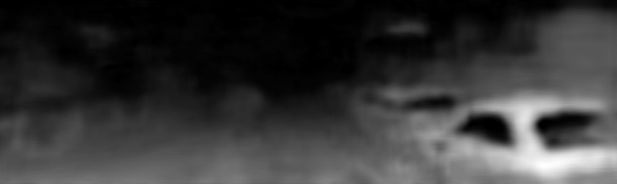}
   \caption{\label{fig:stereo}\textbf{Stereovision test.} Top: right image; middle: ground truth depth map for right image; down: predicted depth map.}
\end{center}
\end{figure*}

\subsection{Stereovision}

In the paper we suggest to try the same framework on a different task, the one of stereovision.
The result, shown in Figure \ref{fig:stereo} without any tuning, is very promising and confirms the generalization ability of our approach.